\documentclass[11pt]{article}

\usepackage[final]{acl}

\usepackage{times}
\usepackage{latexsym}
\usepackage{comment} 
\usepackage{enumitem} 

\usepackage[T1]{fontenc}

\usepackage[utf8]{inputenc}

\usepackage{microtype}

\usepackage{inconsolata}

\usepackage{graphicx}

\usepackage{amsmath}

\usepackage{listings}
\usepackage{booktabs}

\usepackage{amssymb}

\usepackage{tikz}
\usetikzlibrary{shapes, arrows.meta, positioning, decorations.pathmorphing, calc, decorations.pathreplacing}

\DeclareRobustCommand{\gridpos}[1]{%
  \tikz[baseline=0.1em]{
    \draw[gray, thin, rounded corners=0.5pt] (0,0) rectangle (1.2em, 0.8em);
    \ifnum#1=1 \fill[black] (0.2em, 0.6em) circle (0.12em); \fi 
    \ifnum#1=2 \fill[black] (0.6em, 0.6em) circle (0.12em); \fi 
    \ifnum#1=3 \fill[black] (1.0em, 0.6em) circle (0.12em); \fi 
    \ifnum#1=4 \fill[black] (0.2em, 0.2em) circle (0.12em); \fi 
    \ifnum#1=5 \fill[black] (0.6em, 0.2em) circle (0.12em); \fi 
    \ifnum#1=6 \fill[black] (1.0em, 0.2em) circle (0.12em); \fi 
  }%
}

\newcommand{\DatasetSplitRatios}{70/15/15}

\newcommand{\MultimodalProblemCount}{10,000}

\newcommand{\HDSAllCount}{1000}
\newcommand{\TrapsCount}{30}

\newcommand{\HDSProbeCount}{144}

\newcommand{\HDSProbeAccuracyMeanCLabel}{Accuracy (Mean C = 177.4)}

\newcommand{\HDSTestPerpDDThirtyB}{7.02}
\newcommand{\HDSTestPerpOTThirtyB}{7.09}
\newcommand{\HDSTestPerpRCThirtyB}{7.08}

\newcommand{\HDSTestPerpDDSEThirtyB}{0.02}
\newcommand{\HDSTestPerpOTSEThirtyB}{0.01}
\newcommand{\HDSTestPerpRCSEThirtyB}{0.01}

\newcommand{\HDSTestDDDetectionThirtyB}{68\%}

\newcommand{\HDSTestOTDetectionThirtyB}{24\%}

\newcommand{\HDSTestDDDetectionImageThirtyB}{91\%}

\newcommand{\HDSTestOTDetectionImageThirtyB}{16\%}

\newcommand{\ContrastiveCountThirtyB}{144}
\newcommand{\ContrastivePrefRateThirtyB}{100.0\%}
\newcommand{\ContrastivePrefRateSEThirtyB}{0.0\%}
\newcommand{\ContrastiveDeltaThirtyB}{+0.2648}
\newcommand{\ContrastiveDeltaSEThirtyB}{+0.0190}

\newcommand{\ContrastiveDDPrefRateThirtyB}{100.0\%}
\newcommand{\ContrastiveDDPrefRateSEThirtyB}{0.0\%}
\newcommand{\ContrastiveDDDeltaThirtyB}{+0.1209}
\newcommand{\ContrastiveDDDeltaSEThirtyB}{+0.0121}

\newcommand{\ContrastiveOTPrefRateThirtyB}{100.0\%}
\newcommand{\ContrastiveOTPrefRateSEThirtyB}{0.0\%}
\newcommand{\ContrastiveOTDeltaThirtyB}{+0.5538}
\newcommand{\ContrastiveOTDeltaSEThirtyB}{+0.0107}

\newcommand{\ContrastiveRCPrefRateThirtyB}{100.0\%}
\newcommand{\ContrastiveRCPrefRateSEThirtyB}{0.0\%}
\newcommand{\ContrastiveRCDeltaThirtyB}{+0.0931}
\newcommand{\ContrastiveRCDeltaSEThirtyB}{+0.0096}

\newcommand{\ContrastivePrefRateImageThirtyB}{97.9\%}
\newcommand{\ContrastivePrefRateSEImageThirtyB}{1.2\%}
\newcommand{\ContrastiveDeltaImageThirtyB}{+0.2119}
\newcommand{\ContrastiveDeltaSEImageThirtyB}{+0.0179}

\newcommand{\ContrastiveDDPrefRateImageThirtyB}{100.0\%}
\newcommand{\ContrastiveDDPrefRateSEImageThirtyB}{0.0\%}
\newcommand{\ContrastiveDDDeltaImageThirtyB}{+0.0652}
\newcommand{\ContrastiveDDDeltaSEImageThirtyB}{+0.0073}

\newcommand{\ContrastiveOTPrefRateImageThirtyB}{100.0\%}
\newcommand{\ContrastiveOTPrefRateSEImageThirtyB}{0.0\%}
\newcommand{\ContrastiveOTDeltaImageThirtyB}{+0.4856}
\newcommand{\ContrastiveOTDeltaSEImageThirtyB}{+0.0127}

\newcommand{\ContrastiveRCPrefRateImageThirtyB}{93.9\%}
\newcommand{\ContrastiveRCPrefRateSEImageThirtyB}{3.4\%}
\newcommand{\ContrastiveRCDeltaImageThirtyB}{+0.0587}
\newcommand{\ContrastiveRCDeltaSEImageThirtyB}{+0.0091}

\newcommand{\TemplateBalancedAccuracyThirtyB}{49.3\%}
\newcommand{\TemplateBalancedHeuristicMatchThirtyB}{34.0\%}
\newcommand{\TemplateBalancedMeanStdThirtyB}{0.3928}

\newcommand{\TemplateStyleMismatchAccuracyThirtyB}{50.7\%}
\newcommand{\TemplateStyleMismatchHeuristicMatchThirtyB}{31.2\%}
\newcommand{\TemplateStyleMismatchMeanStdThirtyB}{0.7136}

\newcommand{\TemplateBalancedAccuracyImageThirtyB}{46.5\%}
\newcommand{\TemplateBalancedHeuristicMatchImageThirtyB}{33.3\%}
\newcommand{\TemplateBalancedMeanStdImageThirtyB}{0.3821}

\newcommand{\TemplateStyleMismatchAccuracyImageThirtyB}{46.5\%}
\newcommand{\TemplateStyleMismatchHeuristicMatchImageThirtyB}{34.0\%}
\newcommand{\TemplateStyleMismatchMeanStdImageThirtyB}{0.6649}

\newcommand{\TrapsPerpDDThirtyB}{7.42}
\newcommand{\TrapsPerpOTThirtyB}{7.30}
\newcommand{\TrapsPerpRCThirtyB}{7.14}

\newcommand{\TrapsPerpDDSEThirtyB}{0.02}
\newcommand{\TrapsPerpOTSEThirtyB}{0.02}
\newcommand{\TrapsPerpRCSEThirtyB}{0.05}

\newcommand{\HDSTestAccuracyAltThirtyB}{49.3\%}
\newcommand{\HDSTestAccuracyAltSEThirtyB}{4.2\%}
\newcommand{\HDSTestNeutralLossAltThirtyB}{7.5265}

\newcommand{\HDSTestDeltaDDAltThirtyB}{-0.5060}
\newcommand{\HDSTestDeltaOTAltThirtyB}{-0.4398}
\newcommand{\HDSTestDeltaRCAltThirtyB}{-0.4441}

\newcommand{\HDSTestDDTargetSupportAltThirtyB}{27.9\%}
\newcommand{\HDSTestDDTargetSupportAltSEThirtyB}{0.3\%}

\newcommand{\HDSTestOTTargetSupportAltThirtyB}{25.8\%}
\newcommand{\HDSTestOTTargetSupportAltSEThirtyB}{0.4\%}

\newcommand{\HDSTestRCTargetSupportAltThirtyB}{26.5\%}
\newcommand{\HDSTestRCTargetSupportAltSEThirtyB}{0.5\%}

\newcommand{\HDSTestAccuracyImageAltThirtyB}{46.5\%}
\newcommand{\HDSTestAccuracyImageAltSEThirtyB}{4.2\%}
\newcommand{\HDSTestNeutralLossImageAltThirtyB}{4.5022}

\newcommand{\HDSTestDeltaDDImageAltThirtyB}{+0.9156}
\newcommand{\HDSTestDeltaOTImageAltThirtyB}{+0.9776}
\newcommand{\HDSTestDeltaRCImageAltThirtyB}{+1.6518}

\newcommand{\HDSTestDDTargetSupportImageAltThirtyB}{22.3\%}
\newcommand{\HDSTestDDTargetSupportImageAltSEThirtyB}{0.3\%}

\newcommand{\HDSTestOTTargetSupportImageAltThirtyB}{17.8\%}
\newcommand{\HDSTestOTTargetSupportImageAltSEThirtyB}{0.4\%}

\newcommand{\HDSTestRCTargetSupportImageAltThirtyB}{10.5\%}
\newcommand{\HDSTestRCTargetSupportImageAltSEThirtyB}{0.3\%}

\newcommand{\LoRARCTrainNThirtyB}{845}
\newcommand{\LoRARCValNThirtyB}{155}

\newcommand{\LoRADDTrainNThirtyB}{852}
\newcommand{\LoRADDValNThirtyB}{148}

\newcommand{\LoRAOTTrainNThirtyB}{830}
\newcommand{\LoRAOTValNThirtyB}{170}

\newcommand{\NudgeTestCountThirtyB}{144}
\newcommand{\NudgeLoRACountThirtyB}{3}
\newcommand{\NudgeTotalComparisonsThirtyB}{432}

\newcommand{\NudgeTotalFlipsThirtyB}{114}
\newcommand{\NudgeImprovedThirtyB}{1}
\newcommand{\NudgeDegradedThirtyB}{113}
\newcommand{\NudgeDetectionFlipsThirtyB}{313}
\newcommand{\NudgeDegradedCarryDropThirtyB}{0}
\newcommand{\NudgeDegradedPartialProductThirtyB}{12}
\newcommand{\NudgeDegradedMagnitudeSlipThirtyB}{0}

\newcommand{\CorrOTDDThirtyB}{0.0726}
\newcommand{\CorrOTRCThirtyB}{0.1247}
\newcommand{\CorrDDRCThirtyB}{0.1192}

\newcommand{\EffectiveUpdateSameSeedAvgThirtyB}{0.2553}
\newcommand{\EffectiveUpdatePrimaryCrossAvgThirtyB}{0.1055}
\newcommand{\EffectiveUpdateSeedGapThirtyB}{0.1498}

\newcommand{\HDSTestPerpDDTwoThirtyFiveB}{6.38}
\newcommand{\HDSTestPerpOTTwoThirtyFiveB}{7.25}
\newcommand{\HDSTestPerpRCTwoThirtyFiveB}{6.78}

\newcommand{\HDSTestPerpDDSETwoThirtyFiveB}{0.01}
\newcommand{\HDSTestPerpOTSETwoThirtyFiveB}{0.01}
\newcommand{\HDSTestPerpRCSETwoThirtyFiveB}{0.01}

\newcommand{\HDSTestDDDetectionTwoThirtyFiveB}{93\%}

\newcommand{\HDSTestOTDetectionTwoThirtyFiveB}{0\%}

\newcommand{\HDSTestDDDetectionImageTwoThirtyFiveB}{100\%}

\newcommand{\HDSTestOTDetectionImageTwoThirtyFiveB}{0\%}

\newcommand{\ContrastivePrefRateTwoThirtyFiveB}{100.0\%}
\newcommand{\ContrastivePrefRateSETwoThirtyFiveB}{0.0\%}
\newcommand{\ContrastiveDeltaTwoThirtyFiveB}{+0.8804}
\newcommand{\ContrastiveDeltaSETwoThirtyFiveB}{+0.0717}

\newcommand{\ContrastiveDDPrefRateTwoThirtyFiveB}{100.0\%}
\newcommand{\ContrastiveDDPrefRateSETwoThirtyFiveB}{0.0\%}
\newcommand{\ContrastiveDDDeltaTwoThirtyFiveB}{+1.9840}
\newcommand{\ContrastiveDDDeltaSETwoThirtyFiveB}{+0.1096}

\newcommand{\ContrastiveOTPrefRateTwoThirtyFiveB}{100.0\%}
\newcommand{\ContrastiveOTPrefRateSETwoThirtyFiveB}{0.0\%}
\newcommand{\ContrastiveOTDeltaTwoThirtyFiveB}{+0.6141}
\newcommand{\ContrastiveOTDeltaSETwoThirtyFiveB}{+0.0152}

\newcommand{\ContrastiveRCPrefRateTwoThirtyFiveB}{100.0\%}
\newcommand{\ContrastiveRCPrefRateSETwoThirtyFiveB}{0.0\%}
\newcommand{\ContrastiveRCDeltaTwoThirtyFiveB}{+0.1665}
\newcommand{\ContrastiveRCDeltaSETwoThirtyFiveB}{+0.0113}

\newcommand{\ContrastivePrefRateImageTwoThirtyFiveB}{100.0\%}
\newcommand{\ContrastivePrefRateSEImageTwoThirtyFiveB}{0.0\%}
\newcommand{\ContrastiveDeltaImageTwoThirtyFiveB}{+0.2752}
\newcommand{\ContrastiveDeltaSEImageTwoThirtyFiveB}{+0.0205}

\newcommand{\ContrastiveDDPrefRateImageTwoThirtyFiveB}{100.0\%}
\newcommand{\ContrastiveDDPrefRateSEImageTwoThirtyFiveB}{0.0\%}
\newcommand{\ContrastiveDDDeltaImageTwoThirtyFiveB}{+0.0901}
\newcommand{\ContrastiveDDDeltaSEImageTwoThirtyFiveB}{+0.0068}

\newcommand{\ContrastiveOTPrefRateImageTwoThirtyFiveB}{100.0\%}
\newcommand{\ContrastiveOTPrefRateSEImageTwoThirtyFiveB}{0.0\%}
\newcommand{\ContrastiveOTDeltaImageTwoThirtyFiveB}{+0.5918}
\newcommand{\ContrastiveOTDeltaSEImageTwoThirtyFiveB}{+0.0131}

\newcommand{\ContrastiveRCPrefRateImageTwoThirtyFiveB}{100.0\%}
\newcommand{\ContrastiveRCPrefRateSEImageTwoThirtyFiveB}{0.0\%}
\newcommand{\ContrastiveRCDeltaImageTwoThirtyFiveB}{+0.1118}
\newcommand{\ContrastiveRCDeltaSEImageTwoThirtyFiveB}{+0.0100}

\newcommand{\TemplateBalancedAccuracyTwoThirtyFiveB}{57.6\%}
\newcommand{\TemplateBalancedHeuristicMatchTwoThirtyFiveB}{28.5\%}
\newcommand{\TemplateBalancedMeanStdTwoThirtyFiveB}{0.3987}

\newcommand{\TemplateStyleMismatchAccuracyTwoThirtyFiveB}{64.6\%}
\newcommand{\TemplateStyleMismatchHeuristicMatchTwoThirtyFiveB}{30.6\%}
\newcommand{\TemplateStyleMismatchMeanStdTwoThirtyFiveB}{0.7122}

\newcommand{\TemplateBalancedAccuracyImageTwoThirtyFiveB}{53.5\%}
\newcommand{\TemplateBalancedHeuristicMatchImageTwoThirtyFiveB}{30.6\%}
\newcommand{\TemplateBalancedMeanStdImageTwoThirtyFiveB}{0.4516}

\newcommand{\TemplateStyleMismatchAccuracyImageTwoThirtyFiveB}{50.0\%}
\newcommand{\TemplateStyleMismatchHeuristicMatchImageTwoThirtyFiveB}{32.6\%}
\newcommand{\TemplateStyleMismatchMeanStdImageTwoThirtyFiveB}{0.6663}

\newcommand{\HDSTestAccuracyAltTwoThirtyFiveB}{57.6\%}
\newcommand{\HDSTestAccuracyAltSETwoThirtyFiveB}{4.1\%}
\newcommand{\HDSTestNeutralLossAltTwoThirtyFiveB}{4.9058}

\newcommand{\HDSTestDeltaDDAltTwoThirtyFiveB}{+1.4734}
\newcommand{\HDSTestDeltaOTAltTwoThirtyFiveB}{+2.3409}
\newcommand{\HDSTestDeltaRCAltTwoThirtyFiveB}{+1.8791}

\newcommand{\HDSTestDDTargetSupportAltTwoThirtyFiveB}{15.2\%}
\newcommand{\HDSTestDDTargetSupportAltSETwoThirtyFiveB}{0.2\%}

\newcommand{\HDSTestOTTargetSupportAltTwoThirtyFiveB}{5.9\%}
\newcommand{\HDSTestOTTargetSupportAltSETwoThirtyFiveB}{0.3\%}

\newcommand{\HDSTestRCTargetSupportAltTwoThirtyFiveB}{12.4\%}
\newcommand{\HDSTestRCTargetSupportAltSETwoThirtyFiveB}{0.3\%}

\newcommand{\HDSTestAccuracyImageAltTwoThirtyFiveB}{53.5\%}
\newcommand{\HDSTestAccuracyImageAltSETwoThirtyFiveB}{4.2\%}
\newcommand{\HDSTestNeutralLossImageAltTwoThirtyFiveB}{4.1863}

\newcommand{\HDSTestDeltaDDImageAltTwoThirtyFiveB}{+1.4594}
\newcommand{\HDSTestDeltaOTImageAltTwoThirtyFiveB}{+2.2722}
\newcommand{\HDSTestDeltaRCImageAltTwoThirtyFiveB}{+2.2114}

\newcommand{\HDSTestDDTargetSupportImageAltTwoThirtyFiveB}{18.7\%}
\newcommand{\HDSTestDDTargetSupportImageAltSETwoThirtyFiveB}{0.4\%}

\newcommand{\HDSTestOTTargetSupportImageAltTwoThirtyFiveB}{6.8\%}
\newcommand{\HDSTestOTTargetSupportImageAltSETwoThirtyFiveB}{0.3\%}

\newcommand{\HDSTestRCTargetSupportImageAltTwoThirtyFiveB}{8.8\%}
\newcommand{\HDSTestRCTargetSupportImageAltSETwoThirtyFiveB}{0.3\%}

\newcommand{\LoRARCTrainNTwoThirtyFiveB}{845}
\newcommand{\LoRARCValNTwoThirtyFiveB}{155}

\newcommand{\LoRADDTrainNTwoThirtyFiveB}{852}
\newcommand{\LoRADDValNTwoThirtyFiveB}{148}

\newcommand{\LoRAOTTrainNTwoThirtyFiveB}{830}
\newcommand{\LoRAOTValNTwoThirtyFiveB}{170}

\newcommand{\NudgeTestCountTwoThirtyFiveB}{144}
\newcommand{\NudgeLoRACountTwoThirtyFiveB}{3}
\newcommand{\NudgeTotalComparisonsTwoThirtyFiveB}{432}

\newcommand{\NudgeTotalFlipsTwoThirtyFiveB}{121}
\newcommand{\NudgeImprovedTwoThirtyFiveB}{4}
\newcommand{\NudgeDegradedTwoThirtyFiveB}{117}
\newcommand{\NudgeDetectionFlipsTwoThirtyFiveB}{291}
\newcommand{\NudgeDegradedCarryDropTwoThirtyFiveB}{0}
\newcommand{\NudgeDegradedPartialProductTwoThirtyFiveB}{8}
\newcommand{\NudgeDegradedMagnitudeSlipTwoThirtyFiveB}{0}

\newcommand{\CorrOTDDTwoThirtyFiveB}{0.0412}
\newcommand{\CorrOTRCTwoThirtyFiveB}{0.0586}
\newcommand{\CorrDDRCTwoThirtyFiveB}{0.0342}

\newcommand{\EffectiveUpdateSameSeedAvgTwoThirtyFiveB}{0.1476}
\newcommand{\EffectiveUpdatePrimaryCrossAvgTwoThirtyFiveB}{0.0447}
\newcommand{\EffectiveUpdateSeedGapTwoThirtyFiveB}{0.1029}

\providecommand{\HDSProbeCount}{N/A}

\providecommand{\HDSProbeAccuracyMeanCLabel}{Accuracy (Mean C = \HDSProbeMeanCDisplay{})}
\providecommand{\TemplateBalancedMeanStdThirtyB}{N/A}
\providecommand{\TemplateBalancedMeanStdImageThirtyB}{N/A}
\providecommand{\TemplateStyleMismatchMeanStdThirtyB}{N/A}
\providecommand{\TemplateStyleMismatchMeanStdImageThirtyB}{N/A}
\providecommand{\TemplateBalancedMeanStdTwoThirtyFiveB}{N/A}
\providecommand{\TemplateBalancedMeanStdImageTwoThirtyFiveB}{N/A}
\providecommand{\TemplateStyleMismatchMeanStdTwoThirtyFiveB}{N/A}
\providecommand{\TemplateStyleMismatchMeanStdImageTwoThirtyFiveB}{N/A}
\providecommand{\EffectiveUpdateSameSeedAvgThirtyB}{N/A}
\providecommand{\EffectiveUpdatePrimaryCrossAvgThirtyB}{N/A}
\providecommand{\EffectiveUpdateSeedGapThirtyB}{N/A}
\providecommand{\EffectiveUpdateSameSeedAvgTwoThirtyFiveB}{N/A}
\providecommand{\EffectiveUpdatePrimaryCrossAvgTwoThirtyFiveB}{N/A}
\providecommand{\EffectiveUpdateSeedGapTwoThirtyFiveB}{N/A}

\title{
Multiplication in Multimodal LLMs: \\ Computation with Text, Image, and Audio Inputs\thanks{To appear in ACL Findings (2026).}
}

\author{
  Samuel G.\ Balter \\  University of Texas at Austin \\ \texttt{sgb2634@eid.utexas.edu}
        \And Ethan Jerzak  \\ National University of Singapore (NUS) \\ \texttt{phiejj@nus.edu.sg}
       \AND  Connor T.\ Jerzak \\ University of Texas at Austin\\\texttt{connor.jerzak@austin.utexas.edu}
}

\begin{document}
\maketitle
\begin{abstract}
\vspace{-0.2cm}
Multimodal LLMs can accurately perceive numerical content across modalities yet fail to perform exact multi-digit multiplication when the identical underlying arithmetic problem is presented as numerals, number words, images, or in audio form. Because existing benchmarks often lack systematically paired instances across modalities, it remains difficult to compare genuine arithmetic limits within and across model families. We therefore introduce a controlled multimodal multiplication benchmark that factorially varies digit length, digit sparsity, representation (e.g., numerals vs. number words), and modality (text, rendered images, and audio), with paired instances from a reproducible generator. We also define arithmetic load, $C$, as the product of the \textsc{Total} and \textsc{Non-Zero} digit count as a compact, mechanistically motivated proxy for operation count. Across evaluations, accuracy falls sharply as $C$ grows, often nearing zero by $C>$100. Indeed, $C$ remains predictive of performance across modalities and models, with $R^2$ $>$ 0.5, nearing the value from more complex measures of arithmetic load that count the number of intermediate arithmetic steps. A separate perception-versus-computation decomposition shows that multimodal degradation is primarily computational rather than perceptual: on matched perception checks, models are near-perfect ($>$99\%) across modalities even when multiplication accuracy drops substantially. Beyond measuring when models fail, we ask which procedures they are predisposed to follow. We introduce a style-controlled forced-completion loss probe that scores heuristic-specific reasoning prefixes---including columnar multiplication, distributive decomposition, and rounding/compensation. Here, distributive decomposition is favored in both text and vision modalities; heuristic-specific LoRA adapters produce near-orthogonal updates yet degrade accuracy, indicating the base model maintains a well-tuned internal router.
\end{abstract}

\section{Introduction}

Large language models (LLMs) increasingly act as general-purpose assistants, and many real workflows require \emph{exact} arithmetic. Unfortunately, arithmetic competence can be unpredictable: a model that correctly computes 47$\times$36 may fail on a nearby instance like 89$\times$67, despite both being two-digit multiplications. Recent evaluations confirm that multi-digit arithmetic can be brittle in LLMs \citep{mahendra-etal-2025-evaluating}; performance can vary with operand size and surface form in ways that are hard to anticipate \citep{yuan2023llm-arithmetic}. Methods such as chain-of-thought prompting and math-focused pretraining improve performance on broader quantitative reasoning tasks \citep{wei2022chain,lewkowycz2022minerva}, but they do not guarantee the reliable execution of the intermediate calculations those approaches rely on. Moreover, widely used benchmarks (e.g., GSM8K) can entangle language understanding, multi-step reasoning, and computation, making it difficult to attribute errors to \emph{arithmetic} rather than to interpretation or planning \citep{cobbe2021gsm8k}.

At the same time, foundation models are increasingly \emph{multimodal}, with arithmetic questions often arriving as rendered equations in images (e.g., screenshots, slides, scanned homework) or audio traces instead of as text tokens \citep{alayrac2022flamingo,openai2023gpt4v}. This issue becomes more important in agentic workflows: an agent may inspect a screenshot, choose a calculation path, call tools, and verify intermediate states. These steps, however, are most effective when the model captures the mathematical problem underlying the model presentation completely, instead of relying on superficial cues in answering. Recent benchmarks and training efforts in multimodal mathematical reasoning emphasize complex visual contexts and diverse problem types \citep{lu2023mathvista,chen2024mathllava}, but comparatively little work isolates elementary arithmetic under controlled problem structure \emph{and} controlled modal format. This gap matters because representation and perception can amplify small computational weaknesses: the same multiplication might be easy as typed numerals but hard when embedded in a noisy visual channel or reformulated as spoken number words.

We address this gap with controlled experiments on multiplication, a particularly sensitive probe because it requires carry propagation and long-range digit interactions. In the benchmark, we systematically vary (i) operand digit length, (ii) the number of non-zero digits, and (iii) modality (paired text, rendered images, and audio). Through this benchmark, we ask \textit{(1)} How do digit length and algorithmic load (total digits $\times$ non-zero digit count) determine accuracy? \textit{(2)} How much does performance change when the same computation is presented across modalities and representations? \textit{(3)}  Which internal arithmetic heuristics do models prefer, how do these priors shift with operand cues and modality, and do they help explain failure patterns?

Beyond accuracy, we will aim to characterize \emph{why} failures occur by probing strategy preference. Human multiplication draws on multiple heuristics---including columnar procedures, distributive decomposition, and rounding/compensation---whose use depends on problem structure \citep{campbell2005cognitive}. Inspired by probing methods for interpreting neural representations \citep{belinkov2022probing}, we introduce a forced-completion loss probe over heuristic-specific reasoning preambles, using only forward passes. Finally, using low-rank adapters \citep{hu2022lora}, we analyze whether heuristic-specialized updates align in parameter space, providing geometric evidence about whether these heuristics correspond to distinct mechanisms.

Our contributions are:
\begin{itemize}[leftmargin=*]
    \item \textbf{Controlled multimodal benchmark.} We introduce a reproducible multiplication benchmark that factorially varies digit length, digit sparsity, representation, and modality under standardized evaluation across modalities. 
    \item \textbf{An interpretable difficulty measure.} We show that arithmetic load (total digits $\times$ non-zero digits) is a strong predictor of accuracy and helps summarize degradation across modalities.
    \item \textbf{Forced-completion loss-based strategy fingerprinting.} We use a forward-pass loss probe and find model-specific low-loss preferences, with margins shifting with operand cues and modality.
    \item \textbf{Geometric evidence for distinct heuristics.} Heuristic-specific LoRA adapters show separated effective update directions (cosine similarity of $\Delta W = BA$), with stronger orthogonality in 235B than in 30B, suggesting distinct parameter subspaces.
\end{itemize}
    
\begin{figure*}[htb]
\centering
\resizebox{\textwidth}{!}{%
\input{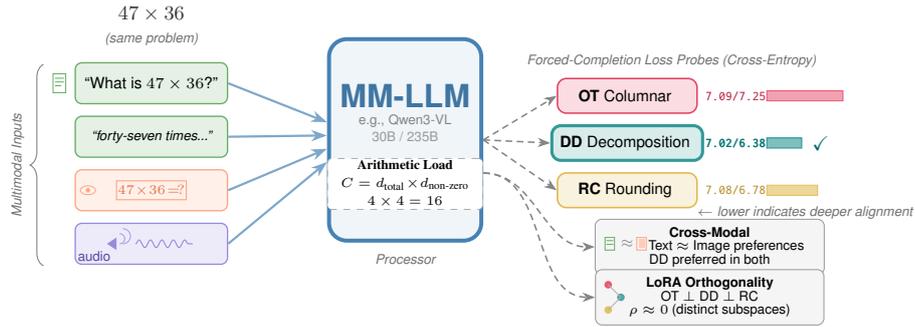}
}
\caption{
  Overview of our arithmetic benchmark and heuristic fingerprinting methodology for multimodal LLMs. In the generator, the same multiplication problem (e.g., $47 \times 36$) is paired across numeric text, rendered numeric image, and audio. 
}
\label{fig:overview}
\end{figure*}

\section{Background}

\subsection{Arithmetic Reliability in LLMs}

To situate our controlled benchmark, we briefly summarize what prior work has established about exact arithmetic in LLMs. Although LLMs can produce fluent arithmetic explanations, controlled evaluations show that exact multi-digit computation can be brittle: accuracy drops quickly as digit length grows and carry interactions accumulate, while small changes in operand configuration can flip correctness even under deterministic decoding \citep{yuan2023llm-arithmetic}. Prompting methods such as chain-of-thought can improve performance on many reasoning benchmarks by eliciting intermediate steps \citep{wei2022chain}, and math-focused pretraining and verification-style training can further raise end-to-end scores on quantitative tasks \citep{lewkowycz2022minerva,cobbe2021gsm8k}. However, these approaches still rely on the model executing intermediate arithmetic correctly \citep{yuan2023llm-arithmetic}. Moreover, widely used datasets such as GSM8K provide end-to-end correctness labels for word problems, which entangle language understanding, planning, and computation in ways that make arithmetic-specific failure modes difficult to isolate \citep{cobbe2021gsm8k}.

\subsection{Representation and Modality Effects}

Beyond operand size and carry interactions, arithmetic success is also sensitive to how a given computation is encoded and presented to the model. As noted, arithmetic questions increasingly arrive in diverse formats---typed numerals, number words, and multimodal inputs such as screenshots or rendered equations---as foundation models extend from text-only to vision-language interfaces \citep{alayrac2022flamingo,openai2023gpt4v}. Recent multimodal benchmarks and instruction-tuned systems increasingly target broad mathematical reasoning in visual contexts \citep{lu2023mathvista,chen2024mathllava}, emphasizing heterogeneous problem types with rich visual scenes (but with few, if any, evaluations that isolate elementary arithmetic under controlled modal variation). As a result, it remains unclear how much multimodal degradation reflects perceptual/representation noise versus limitations in the underlying computation \citep{liu2024ocrbench}, and whether failures transfer systematically across modalities when the arithmetic itself is held fixed \citep{lu2023mathvista,chen2024mathllava}.

\subsection{Arithmetic Heuristics and Strategy Probing}

Even when the input format is held fixed, multiplication can be carried out via multiple procedures, and the choice of strategy affects both intermediate operation counts and characteristic error modes. Work in mathematical cognition emphasizes that multiplication can be performed using multiple solution strategies, with choice depending on operand structure, experience, and context \citep{campbell2005cognitive,dehaene2011number}. We focus on three canonical heuristics that capture some of these distinct computational approaches:
\begin{itemize}[leftmargin=*]
    \item \textbf{Columnar/long multiplication (OT)}: compute digit-by-digit partial products---intermediate products formed by multiplying one operand by a single digit or place-value component of the other, such as $47 \times 6$ and $47 \times 30$ when solving $47 \times 36$---with explicit carry propagation (the standard algorithm in contemporary primary education) \citep{campbell2005cognitive}. OT can be attractive for unstructured or carry-heavy operands because it systematically computes every digit-by-digit partial product.
    \item \textbf{Distributive decomposition (DD)}: rewrite an operand as a sum and add partial products, e.g., $47 \times 36 = 40 \times 36 + 7 \times 36$ \citep{campbell2005cognitive,dehaene2011number}. DD can be attractive when one operand has trailing zeros or other place-value structure because it converts a dense multiplication into a small number of simpler subproducts.
    \item \textbf{Rounding--compensation (RC)}: round to a convenient base and correct, e.g., $49 \times 51 = (50-1)(50+1) = 50^2 - 1$ \citep{campbell2005cognitive}. RC can be attractive when operands are close to round numbers because it converts a dense multiplication into simpler round-base operations plus a small correction.
\end{itemize}
These heuristics imply different computational burdens (number of partial products, carries, and adjustments) and characteristic failure modes when steps are omitted or misapplied \citep{campbell2005cognitive,dehaene2011number}. To uncover which of these procedural paths a given model naturally favors, we will draw on probing methods from the interpretability literature, which provide tools for testing the compatibility between specific computations and a model's internal representations \citep{belinkov2022probing}.

\section{Methodology}

\subsection{Problem Design}

Building on the evidence above that arithmetic failures depend on both operand structure and surface form, we here design a controlled multiplication benchmark that lets us vary these factors independently. We study multiplication in a controlled setting by generating operand pairs from digit templates that vary the number and arrangement of non-zero digits. Concretely, we sample problems from the template family shown in Table~\ref{tab:digit_templates}, which spans fully dense numbers (e.g., VV, VVV) as well as structured sparsity through trailing zeros (e.g., V0, V00, VV0) and non-adjacent non-zeros (e.g., V0V). This design lets us vary arithmetic difficulty without changing the task format, yielding a broad range of carry patterns.

\begin{table}[htb]
    \centering
    \small
    \begin{tabular}{ll}
        \hline
        Template & Description \\
        \hline
        V & single random digit (1--9) \\
        VV & two random digits \\
        VVV & three random digits \\
        V0, V00, VV0 & trailing zeros (fewer operations needed) \\
        V0V & non-adjacent non-zero digits \\
        \hline
    \end{tabular}
    \caption{Digit templates used to vary operand structure.}
    \label{tab:digit_templates}
\end{table}

To quantify difficulty with a single interpretable scalar, we define arithmetic load as
\begin{equation}
    C := \text{Load}(a, b) = d_{\text{total}} \times d_{\text{non-zero}},
    \label{eq:Load}
\end{equation}
where $d_{\text{total}}$ is the total number of digits across both operands and $d_{\text{non-zero}}$ is the count of non-zero digits. For example, 47 $\times$ 36 has $d_{\text{total}}$ = 4 and $d_{\text{non-zero}}$ = 4, yielding load $=$ 16. This load score is a simple, though imperfect, proxy for operation count (see \S\ref{sec:formal-notes}). In our sensitivity checks, it preserves nearly the same explanatory signal as more complex alternatives that explicitly count carry-propagation requirements within each multiplication problem. We therefore use $C$ as the main predictor and reserve carry-aware counts for validation. 

\subsection{Model Configuration}

With the problem distribution fixed, we next specify the model families used for two parts of the study: (i) measuring multimodal arithmetic accuracy at scale and (ii) running token-level probes and adaptations that require access to cross-entropy losses. The broader multimodal scaling analysis evaluates Gemini, Qwen, OpenAI, and xAI model families. The forced-completion, behavioral nudge, and adapter-geometry analyses focus on Qwen3-VL-30B-A3B-Instruct and Qwen3-VL-235B-A22B-Instruct, which provide the token-loss access needed for those experiments. We report the Qwen results side by side (30B/235B) in the probing, while the multimodal accuracy curves summarize the broader evaluation setting.

\subsection{Multimodal Evaluation}

To isolate modality effects, we evaluate the same underlying multiplication items while varying only the input channel. The paired benchmark contains \MultimodalProblemCount{} shared multiplication instances, each rendered as text, an image, and (where possible) audio. The regression figures also include broader representation variants, including alphabetic text and rendered alphabetic prompts. %Regression tables, residual analyses, and predictive summaries are provided in \S\ref{sec:load-fits}.

With the paired multimodal instances now in hand, we next quantify degradation with increasing difficulty by fitting, separately per modality, a logistic regression that predicts correctness in problem $i$ from its arithmetic load:
\begin{equation}
    P(\text{correct}_i) = \sigma(\beta_0 + \beta_1 \cdot \text{Load}_i),
\end{equation}
where $\sigma$ is the logistic function. This formulation highlights arithmetic load as a scalar predictor comparable across problem families.

\paragraph{Error-rate model.} If each primitive digit operation fails independently with probability $p$, then $P(\text{correct}_i) = (1-p)^{N_{\text{ops}}} \approx \exp(-p N_{\text{ops}})$, so log-accuracy is linear in the operation count (coarsely proxied by $C$). We use logistic regression as a convenient monotone fit across modalities; its slope ($\beta_1$) should be read such that higher (less negative) values indicate slower degradation as arithmetic load increases, and its intercept ($\beta_0$) such that a higher value indicates stronger baseline performance (see Table \ref{Tab:BigComparison}). A log link on $P(\text{correct}_i)$ matches the independent-error approximation; other monotone links (including log-log) could be reasonable under different parameterizations of $N_{\text{ops}}$ (see \S\ref{sec:formal-notes}).

\subsection{Heuristic Fingerprinting}

Accuracy-versus-load curves tell us \emph{when} models break down, but they do not by themselves reveal \emph{which} intermediate procedure the model is predisposed to follow on a given instance. This is important because it will give insight into whether the observed failures arise from a uniform loss of arithmetic competence or from shifts in the model's preferred computation strategy across problems and modalities. Indeed, DD can be efficient for clean place-value structures, such as $47\times 60$; RC can be efficient for near-base problems, such as $49\times 51$; OT can be helpful for unstructured operands. 

Thus, two mathematically similar problems can become easier or harder depending on the heuristic chosen. The fingerprinting probe is designed to measure these compatibility differences across modalities in a controlled way.

To probe which procedural strategy a model finds most ``natural'' for a given multiplication instance \citep{belinkov2022probing}, we introduce a perplexity-style loss probe implemented with Qwen3-VL-30B-A3B-Instruct and Qwen3-VL-235B-A22B. We infer heuristic preference from relative loss under forced continuation (token-level cross-entropy), enabling large-scale preference measurement without confounds from sampling, verbosity, or post-hoc answer formatting.

For each problem $(a,b)$, we construct a small bank of heuristic-conditioned assistant continuations, each describing a specific computation style in short, operand-free language so that the same template bank can be applied to matched text and image presentations:
\begin{itemize}[leftmargin=*]
    \item \textbf{OT}: ``Columnar: start with the ones digits [...]''
    \item \textbf{DD}: ``Decomposition: split one factor into place-value parts [...]''
    \item \textbf{RC}: ``Round \& adjust: use a nearby round base, then compensate [...]''
\end{itemize}
To ensure that loss differences reflect genuine heuristic alignment, the probe uses a balanced bank of short, stylistically matched paraphrases per heuristic, together with a neutral baseline preamble, so that we can report relative preferences as $\Delta$loss rather than raw loss. We score each heuristic by the length-normalized cross-entropy of its continuation tokens under the same question context (forced completion), averaging across the active template bank, and we interpret \emph{lower} loss (or more negative $\Delta$loss relative to the neutral prompt) as a deeper alignment with that heuristic's initial reasoning trajectory. We treat the minimum-loss aggregated template bank as a per-instance, loss-based label and analyze how both labels and margins shift with operand cues and modality;
we compute loss only on the heuristic preamble/continuation tokens.

\paragraph{Likelihood-ratio view.} Framing the probe as a likelihood ratio will provide a formal statistical justification for interpreting cross-entropy differences as the model's explicit algorithmic preference. Let $\ell(h)$ be the length-normalized cross-entropy over the heuristic template continuation tokens under the same question context (forced completion), and let $\ell_0$ be the neutral baseline. If templates are token-length matched with length $T$, then $-T \Delta \ell(h)$ equals the log-likelihood ratio between candidate templates under the same model and context. Then, choosing the minimum loss heuristic is the maximum-likelihood choice under equal priors. When lengths differ, $\Delta \ell$ should be read as a normalized naturalness score (see \S\ref{sec:formal-notes}). This setup enables cross-modal heuristic comparison.

We evaluate the probe on two components of our benchmark: first, the Heuristic-Disagreement Set (HDS), which contains \HDSAllCount{} problems intentionally chosen so that their operand configurations distinctively favor one of the three target heuristics, such as near-base numbers where RC is attractive. Items are labeled by an OT/DD/RC cost model. Each item stores both a \emph{design family} and a \emph{target heuristic}, defined as the minimum-cost heuristic under that cost model. We enforce uniqueness up to commutativity and split HDS into train/validation/test using \DatasetSplitRatios{}. 
We define simple cost models for OT/DD/RC (see \S\ref{sec:formal-notes}). We keep only instances where the minimum-cost heuristic is uniquely separated from the runner-up by a margin. Table~\ref{tab:hds-snippet} shows a small snippet of HDS entries.

\begin{table}[htb]
  \centering
  \small
  \begin{tabular}{llll}
    \hline
    ID & Problem & Target & Category \\
    \hline
    hds\_000 & $49 \times 51 = 2499$ & RC & \texttt{near\_50\_sym} \\
    hds\_002 & $99 \times 101 = 9999$ & RC & \texttt{near\_100\_sym} \\
    hds\_009 & $47 \times 60 = 2820$ & DD & \texttt{zero\_factor} \\
    hds\_014 & $37 \times 100 = 3700$ & DD & \texttt{hundred\_factor} \\
    hds\_018 & $87 \times 96 = 8352$ & OT & \texttt{carry\_heavy} \\
    hds\_019 & $79 \times 68 = 5372$ & OT & \texttt{carry\_heavy} \\
    \hline
  \end{tabular}
  \caption{
    Snippet of HDS examples showing operands, target heuristics, and category labels. Full data are available at \url{https://huggingface.co/datasets/cjerzak/MultimodalMathBenchmarks}.
  }
  \label{tab:hds-snippet}
\end{table}

Separately, we include Adversarial Traps. These traps deliberately make a familiar cue unreliable: for example, an anti-rounding trap makes RC look tempting while an offset creates an extra correction burden; a missing-term trap tests whether a decomposition trace drops a required partial product. Performance here tests whether heuristic preferences remain robust to misleading cues.

\subsection{Adapter Weight Orthogonality Analysis}

Because loss-based preferences could, in principle, reflect trace-format effects, we additionally connect differences in strategy to the parameter space. To test whether different heuristics correspond to meaningfully distinct computational mechanisms, we train separate LoRA adapters \citep{hu2022lora} for Qwen3-VL-30B and Qwen3-VL-235B using synthetic reasoning traces generated with no overlap with the held-out test problems. For each model, we train three heuristic adapters (RC, DD, OT) plus a STYLE control adapter trained on generic reasoning-trace formatting without heuristic-specific intermediate arithmetic. The three heuristic adapters drive the main nudge and pairwise geometry analyses, while STYLE serves as a control for whether any degradation is heuristic-specific or a broader trace-steering effect. We train the heuristic adapters with early stopping on validation loss\footnote{For 30B: RC $n=\LoRARCTrainNThirtyB$ train, $\LoRARCValNThirtyB$ val; DD $n=\LoRADDTrainNThirtyB$ train, $\LoRADDValNThirtyB$ val; OT $n=\LoRAOTTrainNThirtyB$ train, $\LoRAOTValNThirtyB$ val; for 235B: RC $n=\LoRARCTrainNTwoThirtyFiveB$ train, $\LoRARCValNTwoThirtyFiveB$ val; DD $n=\LoRADDTrainNTwoThirtyFiveB$ train, $\LoRADDValNTwoThirtyFiveB$ val; OT $n=\LoRAOTTrainNTwoThirtyFiveB$ train, $\LoRAOTValNTwoThirtyFiveB$ val.}, keeping the base model fixed and learning only low-rank updates.

For each adapter, we compute the effective update $\Delta W_h = B_h A_h$ across all LoRA modules and flatten $\Delta W_h$ into a vector. We then compute the pairwise cosine similarity between the update vectors of any two heuristic adapters $h_1$ and $h_2$:
\begin{equation}
    \text{sim}(h_1, h_2) = \frac{\Delta W_{h_1} \cdot \Delta W_{h_2}}{\|\Delta W_{h_1}\| \|\Delta W_{h_2}\|}.
\end{equation}
As $\Delta W$ is invariant to the LoRA factorization, $\text{sim}(h_1, h_2)$ provides an invariant proxy for separability that complements behavioral probing, linking heuristic behavior to parameter-space structure.

\section{Results}

With the benchmark, modalities, and probing setup defined, we now report empirical results and use a consistent notation when comparing models. Unless otherwise noted, we report paired Qwen3-VL values as 30B / 235B, each with standard errors ($\pm$SE) across problem instances.

\subsection{Algorithmic Load Degrades Accuracy Across Modalities}

We begin by establishing a baseline for how structural difficulty affects performance, addressing our first research question on algorithmic load. 

\begin{figure*}[ht]
  \centering
  % row 1 
  \includegraphics[width=0.68\columnwidth]{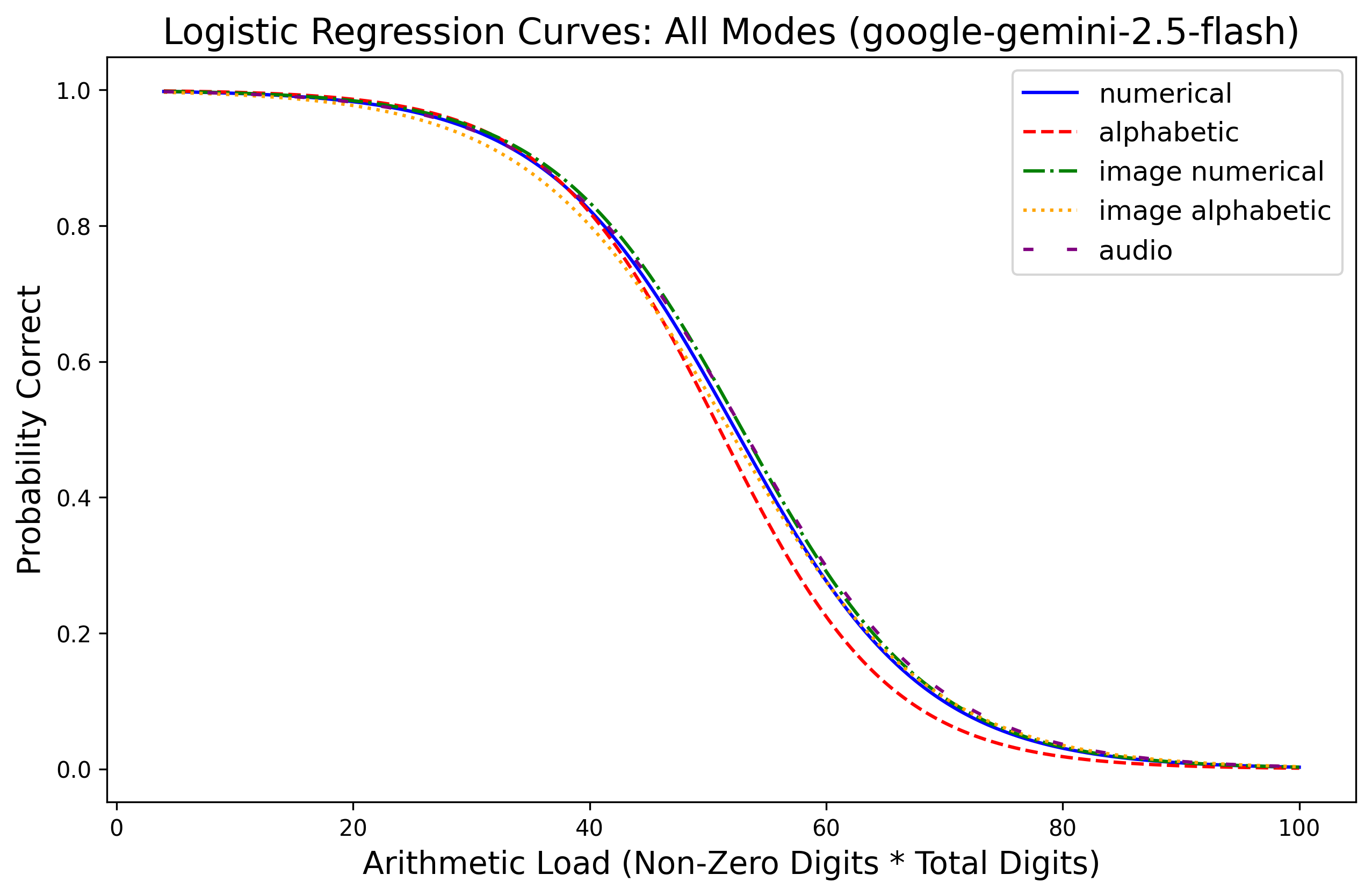}
    \vspace{0.6em}
  \includegraphics[width=0.68\columnwidth]{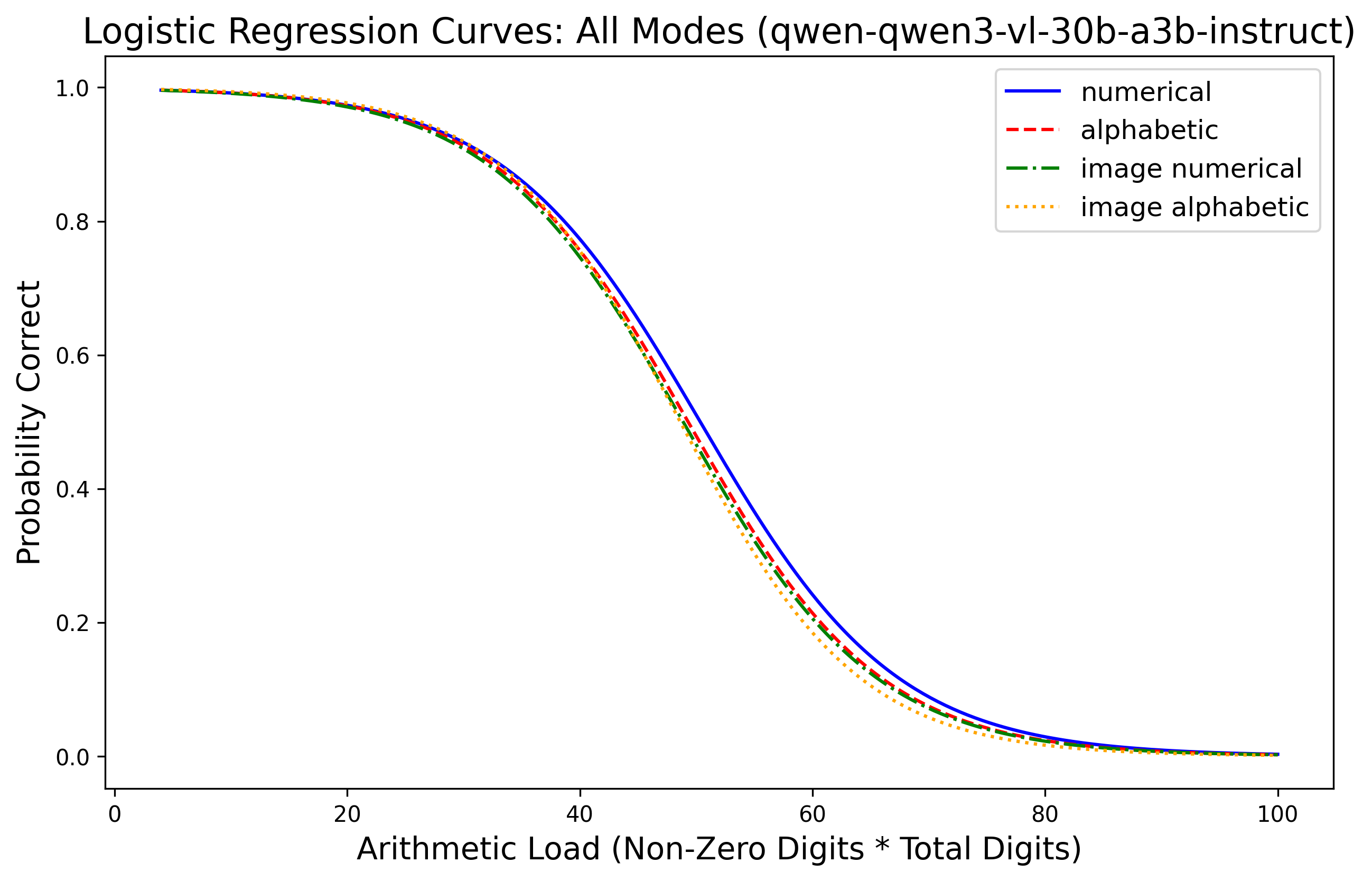}
  \vspace{0.6em}
  \includegraphics[width=0.68\columnwidth]{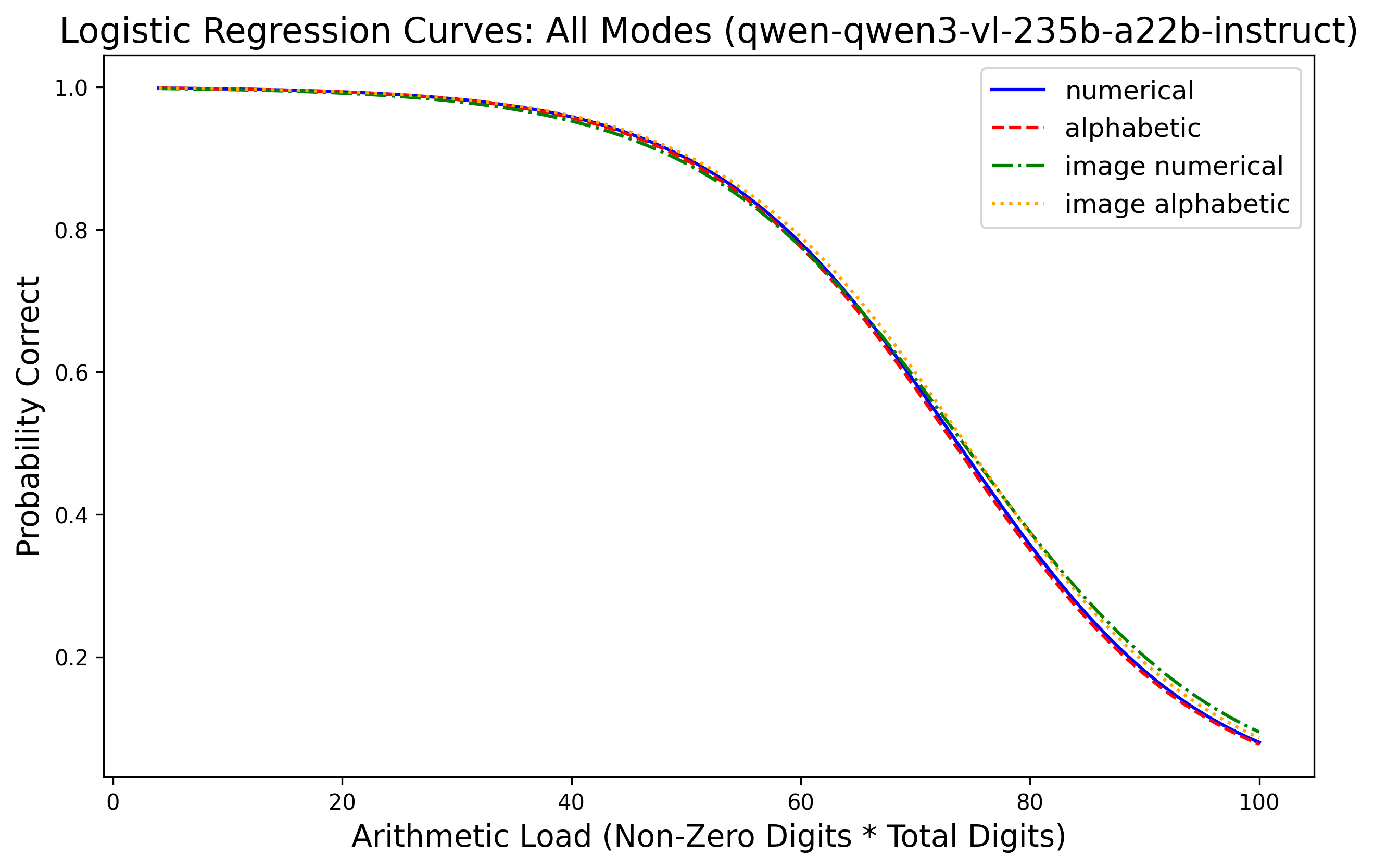}
  % row 2
    \includegraphics[width=0.68\columnwidth]{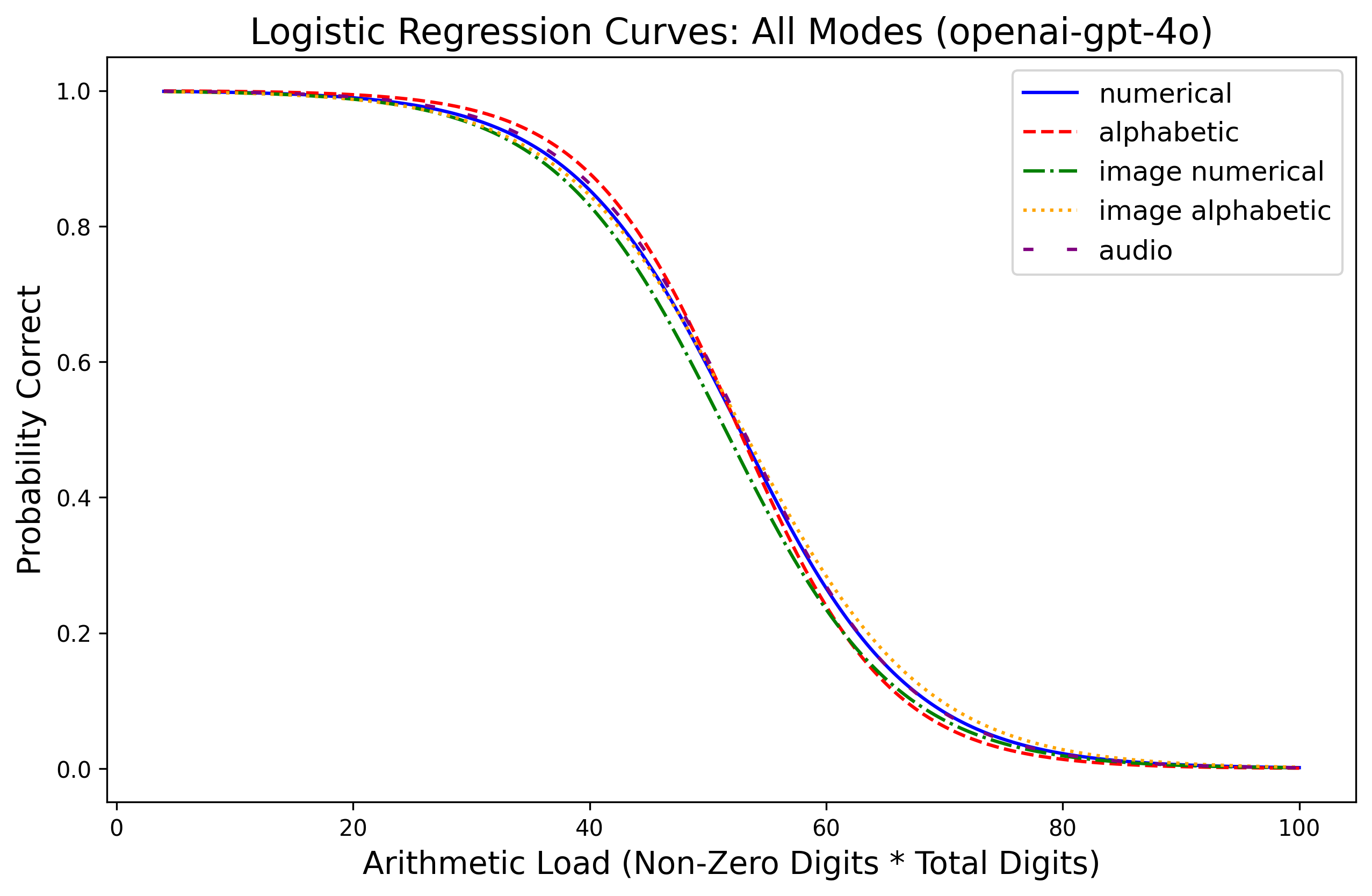}
    \vspace{0.6em}
  \includegraphics[width=0.68\columnwidth]{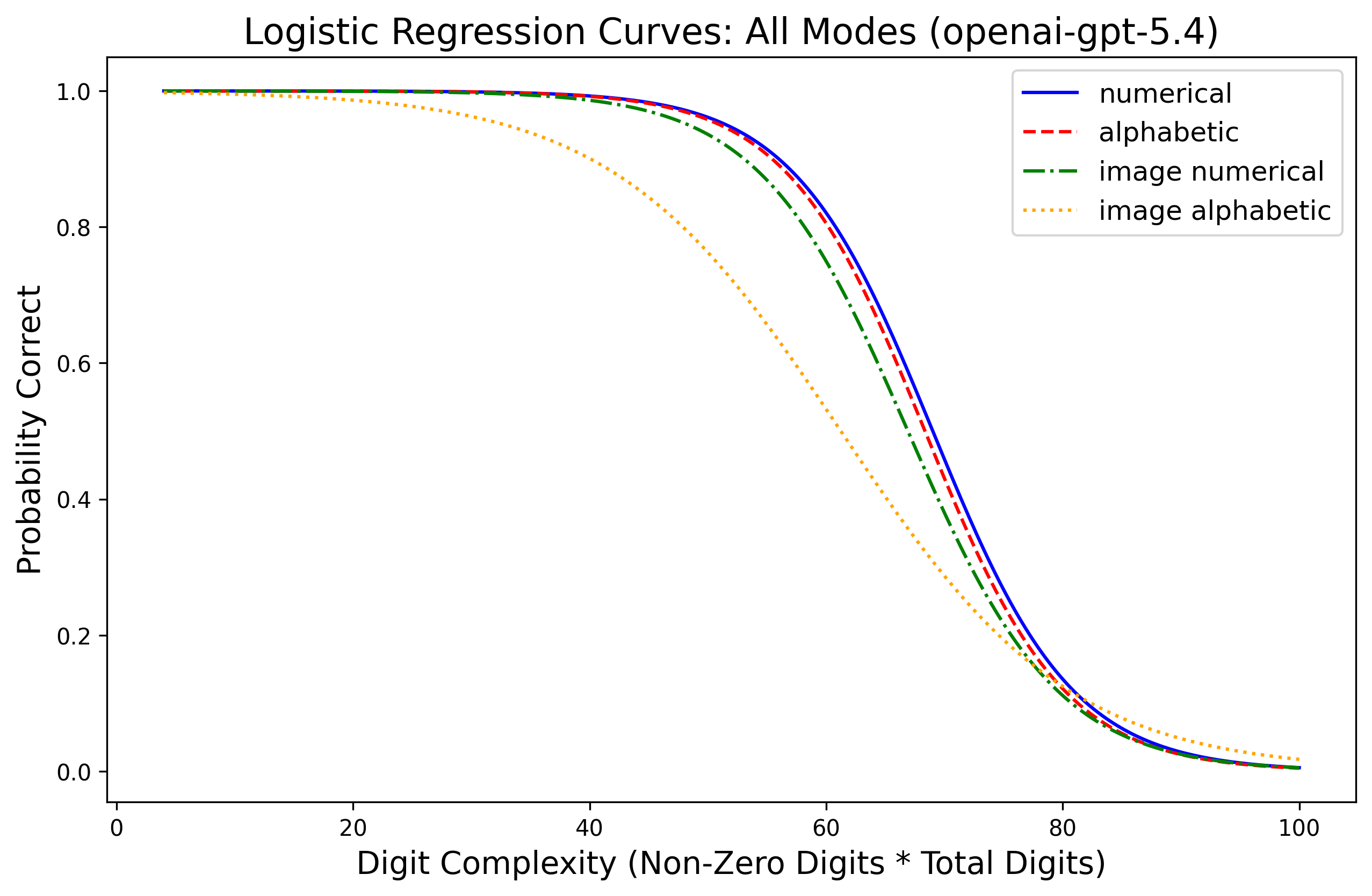}
    \includegraphics[width=0.68\columnwidth]{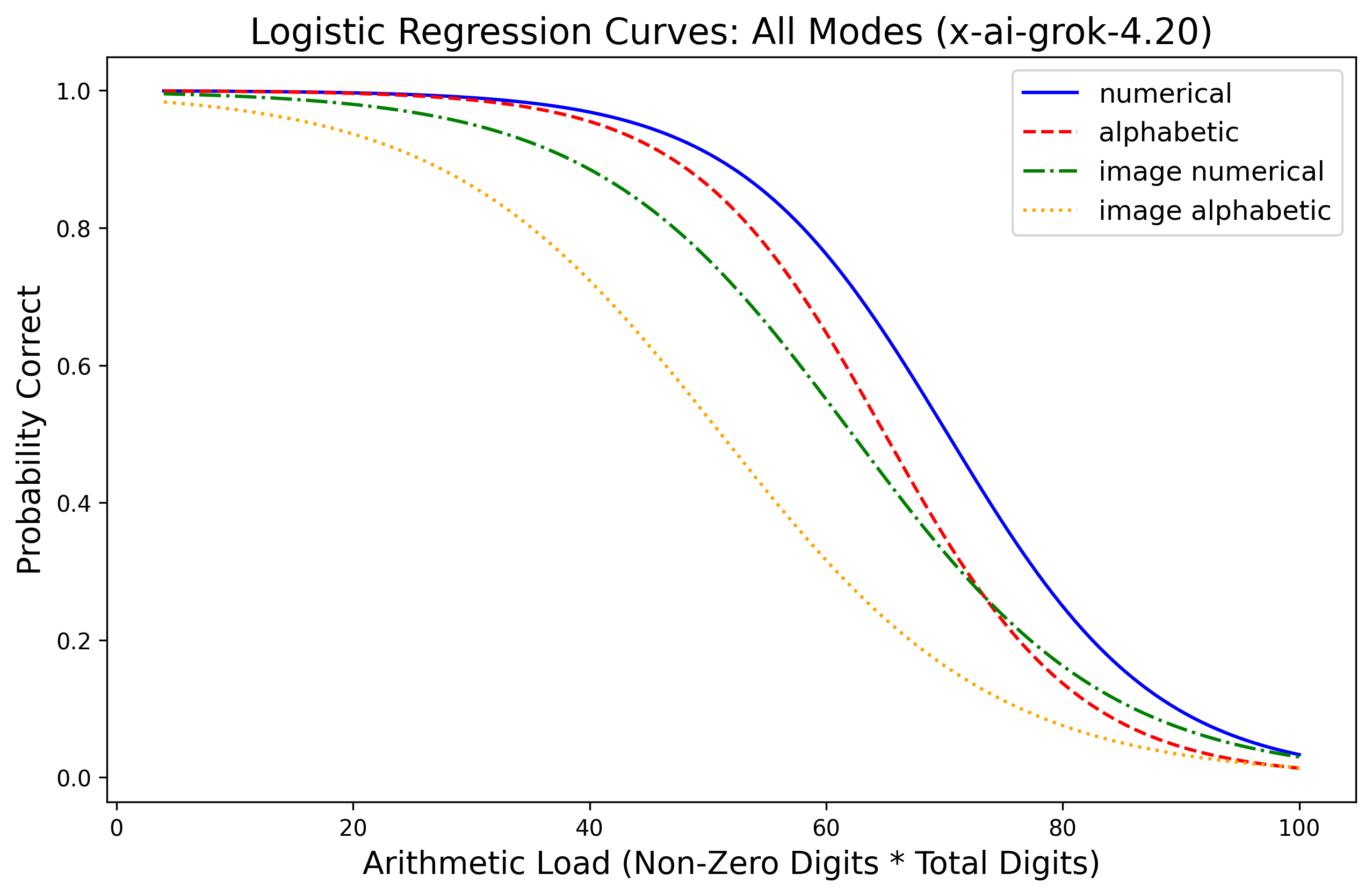}
  \caption{
Probability of correct answer as a function of arithmetic load $C$ (total digits $\times$ non-zero digits) across input modalities for Gemini 2.5 Flash (\gridpos{1}), Qwen3-VL-30B (\gridpos{2}), and Qwen3-VL-235B (\gridpos{3}), with OpenAI's 4o (\gridpos{4}), 5.4 (\gridpos{5}), and xAI's Grok 4.20 (\gridpos{6}) models. Curves show fitted logistic regression models for numeric/alphabetic text, image renderings of those prompts, and audio (where possible).\textit{Note:} Gemini 3.1 Pro performed flawlessly across all modalities within the range of $C$ in the original evaluation suite (1-100), but also took over an hour per modality (as opposed to minutes for other models) and used far more output tokens than other models (by a factor of at least 10). Further trials were therefore run over a larger range of $C$ (1-400). In this extended analysis, Gemini 3.1 Pro began to degrade at a $C$ value of around 360, which occurs, for example, in a problem like:
$
\texttt{1632178320} \times \texttt{5683473970 = ?}. 
$
  }
  \label{fig:Load}
\end{figure*}

Figure~\ref{fig:Load} shows logistic regression curves for the probability of a correct answer versus arithmetic load across modalities. Accuracy declines monotonically with load across modalities, but the ordering depends on representation and model. Overall, text is usually the strongest; audio and image variants tend to lag: curves approach near-zero accuracy at the highest observed difficulty levels. Non-text conditions often have lower intercepts, though fitted slopes are not consistently steeper. Some models (e.g., Gemini 2.5 Flash) show similar scaling performance with $C$ across modalities, while others show greater apparent sensitivity (such as GPT 5.4 or Grok 4.20).

The fitted coefficients in Table \ref{Tab:BigComparison} make these visual trends more explicit. Across model--modality pairs, arithmetic load remains a strong one-dimensional summary of performance: most fits explain roughly half or more of the outcome variation, with $R^2$ values commonly above 0.5. 

The 50\% load threshold is especially useful for comparing models because it combines the intercept and slope into a single difficulty scale. Gemini 2.5 Flash, Qwen3-VL-30B, and GPT-4o all tend to cross 50\% predicted accuracy near $C \approx 50$--$54$, whereas Qwen3-VL-235B remains substantially more robust, with 50\% thresholds near $C \approx 74$--$75$ across its text and image conditions. GPT-5.4 and Grok 4.20 occupy a higher-performing but more modality-sensitive regime: both retain relatively large 50\% thresholds in numerical text or numerical-image settings, but their alphabetic-image conditions drop noticeably.

The table also shows why slopes alone should not be over-interpreted. Some models with steep negative slopes, such as GPT-5.4, nevertheless perform well over a broad range because they begin from a very high intercept, while flatter slopes paired with lower intercepts can still yield lower accuracy at moderate loads. Finally, the modality effects are more often expressed as shifts in intercept or 50\% threshold than as uniformly steeper degradation: numerical images often track numerical text closely, alphabetic images are the most consistent weak point, and audio does not show a uniform penalty in the controlled conditions where it was evaluated. Taken together, these summaries support the main claim that arithmetic load is the dominant axis of degradation, while modality and representation mainly modulate where along that axis each model begins to fail.

\begin{table*}[htb]
\centering \small
\begin{tabular}{llrrrrr}
\hline\hline
\textbf{Model} & \textbf{Modality} & $\beta_0$ & $\beta_1$ & 50\% $C$ & $R^2$ \\
\hline
google-gemini-2.5-flash & numerical & 6.7901 & -0.1271 & 53.4081 & 0.5823 \\
google-gemini-2.5-flash & alphabetic & 6.8299 & -0.1299 & 52.5710 & 0.5944 \\
google-gemini-2.5-flash & image-numerical & 6.4986 & -0.1230 & 52.8540 & 0.5785 \\
google-gemini-2.5-flash & image-alphabetic & 6.5135 & -0.1238 & 52.6159 & 0.5710 \\
google-gemini-2.5-flash & audio & 6.7319 & -0.1246 & 54.0088 & 0.5857 \\
qwen-qwen3-vl-30b-a3b-instruct & numerical & 5.9834 & -0.1167 & 51.2903 & 0.5560 \\
qwen-qwen3-vl-30b-a3b-instruct & alphabetic & 6.3232 & -0.1273 & 49.6883 & 0.5604 \\
qwen-qwen3-vl-30b-a3b-instruct & image-numerical & 5.9858 & -0.1203 & 49.7559 & 0.5364 \\
qwen-qwen3-vl-30b-a3b-instruct & image-alphabetic & 6.3750 & -0.1286 & 49.5609 & 0.5884 \\
qwen-qwen3-vl-235b-a22b-instruct & numerical & 7.0110 & -0.0945 & 74.1790 & 0.3812 \\
qwen-qwen3-vl-235b-a22b-instruct & alphabetic & 7.0143 & -0.0942 & 74.4540 & 0.3693 \\
qwen-qwen3-vl-235b-a22b-instruct & image-numerical & 6.6806 & -0.0889 & 75.1491 & 0.3451 \\
qwen-qwen3-vl-235b-a22b-instruct & image-alphabetic & 6.9417 & -0.0934 & 74.3443 & 0.3753 \\
openai-gpt-5.4 & numerical & 11.0759 & -0.1614 & 68.6056 & 0.6405 \\
openai-gpt-5.4 & alphabetic & 11.2170 & -0.1651 & 67.9324 & 0.6565 \\
openai-gpt-5.4 & image-numerical & 10.5220 & -0.1572 & 66.9180 & 0.6212 \\
openai-gpt-5.4 & image-alphabetic & 6.4656 & -0.1067 & 60.6038 & 0.6182 \\
openai-gpt-4o & numerical & 7.4068 & -0.1398 & 52.9859 & 0.6320 \\
openai-gpt-4o & alphabetic & 8.5268 & -0.1611 & 52.9296 & 0.6883 \\
openai-gpt-4o & image-numerical & 6.9210 & -0.1341 & 51.6105 & 0.6438 \\
openai-gpt-4o & image-alphabetic & 7.1114 & -0.1356 & 52.4258 & 0.6301 \\
openai-gpt-4o-audio-preview & audio & 7.3132 & -0.1378 & 53.0738 & 0.6208 \\
x-ai-grok-4.20 & numerical & 8.1114 & -0.1160 & 69.9351 & 0.5236 \\
x-ai-grok-4.20 & alphabetic & 8.0283 & -0.1230 & 65.2729 & 0.5024 \\
x-ai-grok-4.20 & image-numerical & 5.5582 & -0.0898 & 61.8681 & 0.5078 \\
x-ai-grok-4.20 & image-alphabetic & 4.4572 & -0.0877 & 50.8375 & 0.4071 \\
\hline\hline
\end{tabular}
\caption{
Results across models and modalities. Higher $\beta_0$ indicates stronger baseline performance; higher (less negative) $\beta_1$ indicates slower degradation as arithmetic load increases; 50\% $C$ indicates the load at which predicted accuracy is reduced to 50\% (higher is better). The $R^2$ column indicates the fraction of outcome variability explained by the fit. For an interactive leaderboard, see:
\url{https://neuristemic.ai/multiplication-in-multimodal-llms/}.
}\label{Tab:BigComparison}
\end{table*}

\subsection{Decomposition Preference?}

Having established that arithmetic load predicts degradation across modalities, we next ask whether models exhibit preferences for certain arithmetic procedures that might help explain these failures. To test whether LLMs tend to prefer any single heuristic, we compare how well heuristic-specific preambles match the model's continuation distribution via a forced-completion loss probe.

Table~\ref{tab:modality-comparison-alt} reports mean $\Delta$loss relative to the neutral baseline on a shared held-out split (values $<$ 0 indicate lower forced-completion loss than neutral). Here, target support is the normalized $\exp(-\ell)$ mass assigned to the target heuristic among DD, RC, OT, and the neutral baseline. Because the neutral continuation is included as a competitor, we do not force a heuristic winner when all heuristic cues score worse than neutral.

For both 30B and 235B, DD attains the lowest $\Delta$loss in both text and image modalities. For 30B, DD attains the most negative $\Delta$loss in text and the smallest positive $\Delta$loss in image (\HDSTestDeltaDDAltThirtyB{} / \HDSTestDeltaDDImageAltThirtyB{}), with RC and OT less favorable. For 235B, DD again shows the smallest (least positive) $\Delta$loss in both modalities (\HDSTestDeltaDDAltTwoThirtyFiveB{} / \HDSTestDeltaDDImageAltTwoThirtyFiveB{}), while OT is the least compatible continuation (\HDSTestDeltaOTAltTwoThirtyFiveB{} / \HDSTestDeltaOTImageAltTwoThirtyFiveB{}). The support statistics are likewise DD-dominant across both sizes: 30B places among the largest resolved-probe target support on DD-targeted items (\HDSTestDDTargetSupportAltThirtyB{} $\pm$ \HDSTestDDTargetSupportAltSEThirtyB{} / \HDSTestDDTargetSupportImageAltThirtyB{} $\pm$ \HDSTestDDTargetSupportImageAltSEThirtyB{}), as does 235B (\HDSTestDDTargetSupportAltTwoThirtyFiveB{}$\pm$\HDSTestDDTargetSupportAltSETwoThirtyFiveB{} / \HDSTestDDTargetSupportImageAltTwoThirtyFiveB{}$\pm$\HDSTestDDTargetSupportImageAltSETwoThirtyFiveB{}). 
Raw cross-entropy losses with SE and Traps comparisons are reported in \S\ref{sec:raw-loss}.

  \begin{table*}[ht]
    \centering\small
    \begin{tabular}{llcc}
      \hline
      \textbf{Metric} & \textbf{Model} & \textbf{Text} & \textbf{Image} \\
      \hline
      \HDSProbeAccuracyMeanCLabel{} & 30B & \HDSTestAccuracyAltThirtyB{} $\pm$ \HDSTestAccuracyAltSEThirtyB{} & \HDSTestAccuracyImageAltThirtyB{} $\pm$ \HDSTestAccuracyImageAltSEThirtyB{} \\
      & 235B & \HDSTestAccuracyAltTwoThirtyFiveB{} $\pm$ \HDSTestAccuracyAltSETwoThirtyFiveB{} & \HDSTestAccuracyImageAltTwoThirtyFiveB{} $\pm$ \HDSTestAccuracyImageAltSETwoThirtyFiveB{} \\
      \hline
      \multicolumn{4}{l}{\textit{Neutral baseline loss:}} \\
      \quad Neutral baseline & 30B & \HDSTestNeutralLossAltThirtyB{} & \HDSTestNeutralLossImageAltThirtyB{} \\
      & 235B & \HDSTestNeutralLossAltTwoThirtyFiveB{} & \HDSTestNeutralLossImageAltTwoThirtyFiveB{} \\
      \hline
      \multicolumn{4}{l}{\textit{$\Delta$loss vs.\ neutral under forced completion (lower = deeper alignment):}} \\
      \quad DD (Decomposition) & 30B & \HDSTestDeltaDDAltThirtyB{} & \HDSTestDeltaDDImageAltThirtyB{} \\
      & 235B & \HDSTestDeltaDDAltTwoThirtyFiveB{} & \HDSTestDeltaDDImageAltTwoThirtyFiveB{} \\
      \quad RC (Rounding) & 30B & \HDSTestDeltaRCAltThirtyB{} & \HDSTestDeltaRCImageAltThirtyB{} \\
      & 235B & \HDSTestDeltaRCAltTwoThirtyFiveB{} & \HDSTestDeltaRCImageAltTwoThirtyFiveB{} \\
      \quad OT (Columnar) & 30B & \HDSTestDeltaOTAltThirtyB{} & \HDSTestDeltaOTImageAltThirtyB{} \\
      & 235B & \HDSTestDeltaOTAltTwoThirtyFiveB{} & \HDSTestDeltaOTImageAltTwoThirtyFiveB{} \\
      \hline
      \multicolumn{4}{l}{\textit{Target support on resolved target problems:}} \\
      \quad DD preference share (on DD-targeted items) & 30B & \HDSTestDDTargetSupportAltThirtyB{} $\pm$ \HDSTestDDTargetSupportAltSEThirtyB{} & \HDSTestDDTargetSupportImageAltThirtyB{} $\pm$ \HDSTestDDTargetSupportImageAltSEThirtyB{} \\
      & 235B & \HDSTestDDTargetSupportAltTwoThirtyFiveB{} $\pm$ \HDSTestDDTargetSupportAltSETwoThirtyFiveB{} & \HDSTestDDTargetSupportImageAltTwoThirtyFiveB{} $\pm$ \HDSTestDDTargetSupportImageAltSETwoThirtyFiveB{} \\
      \quad RC preference share (on RC-targeted items) & 30B & \HDSTestRCTargetSupportAltThirtyB{} $\pm$ \HDSTestRCTargetSupportAltSEThirtyB{} & \HDSTestRCTargetSupportImageAltThirtyB{} $\pm$ \HDSTestRCTargetSupportImageAltSEThirtyB{} \\
      & 235B & \HDSTestRCTargetSupportAltTwoThirtyFiveB{} $\pm$ \HDSTestRCTargetSupportAltSETwoThirtyFiveB{} & \HDSTestRCTargetSupportImageAltTwoThirtyFiveB{} $\pm$ \HDSTestRCTargetSupportImageAltSETwoThirtyFiveB{} \\
      \quad OT preference share (on OT-targeted items) & 30B & \HDSTestOTTargetSupportAltThirtyB{} $\pm$ \HDSTestOTTargetSupportAltSEThirtyB{} & \HDSTestOTTargetSupportImageAltThirtyB{} $\pm$ \HDSTestOTTargetSupportImageAltSEThirtyB{} \\
      & 235B & \HDSTestOTTargetSupportAltTwoThirtyFiveB{} $\pm$ \HDSTestOTTargetSupportAltSETwoThirtyFiveB{} & \HDSTestOTTargetSupportImageAltTwoThirtyFiveB{} $\pm$ \HDSTestOTTargetSupportImageAltSETwoThirtyFiveB{} \\
      \hline
    \end{tabular}
    \caption{Modality comparison on a shared held-out probe split ($n=\HDSProbeCount$) for Qwen3-VL-30B and Qwen3-VL-235B. Raw cross-entropy losses are reported in \S\ref{sec:raw-loss}.
    }
    \label{tab:modality-comparison-alt}
  \end{table*}

  %\label{fig:reversal}

A style-shift ablation suggests that the loss-based fingerprint is not just memorizing familiar wording. If the probe were mostly doing surface-form matching, then replacing the balanced template bank with style-shift paraphrases should broadly collapse the aggregate design-family match rate. Yet, heuristic-match moves by only a few points in every setting: \TemplateBalancedHeuristicMatchThirtyB{} to \TemplateStyleMismatchHeuristicMatchThirtyB{} in 30B text, \TemplateBalancedHeuristicMatchImageThirtyB{} to \TemplateStyleMismatchHeuristicMatchImageThirtyB{} in 30B image, \TemplateBalancedHeuristicMatchTwoThirtyFiveB{} to \TemplateStyleMismatchHeuristicMatchTwoThirtyFiveB{} in 235B text, and \TemplateBalancedHeuristicMatchImageTwoThirtyFiveB{} to   \TemplateStyleMismatchHeuristicMatchImageTwoThirtyFiveB{} in 235B image. Accuracy likewise stays in the same broad range rather than collapsing (\TemplateBalancedAccuracyThirtyB{} to \TemplateStyleMismatchAccuracyThirtyB{}, \TemplateBalancedAccuracyImageThirtyB{} to   \TemplateStyleMismatchAccuracyImageThirtyB{}, \TemplateBalancedAccuracyTwoThirtyFiveB{} to   \TemplateStyleMismatchAccuracyTwoThirtyFiveB{}, and \TemplateBalancedAccuracyImageTwoThirtyFiveB{} to \TemplateStyleMismatchAccuracyImageTwoThirtyFiveB{}). The clearest effect of the mismatch is increased noise: the mean within-problem standard deviation across paraphrase losses rises from \TemplateBalancedMeanStdThirtyB{} to \TemplateStyleMismatchMeanStdThirtyB{}, from   \TemplateBalancedMeanStdImageThirtyB{} to \TemplateStyleMismatchMeanStdImageThirtyB{}, from   \TemplateBalancedMeanStdTwoThirtyFiveB{} to \TemplateStyleMismatchMeanStdTwoThirtyFiveB{}, and from   \TemplateBalancedMeanStdImageTwoThirtyFiveB{} to \TemplateStyleMismatchMeanStdImageTwoThirtyFiveB{}   (Table~\ref{tab:template-variability}). In that sense, changing template style acts more like changing an accent than changing the underlying arithmetic cue: the readout gets noisier, but it usually still points to the same family. That said, some family-specific rates are reweighted, especially in 30B image, where DD-target detection falls.
Overall, the ablation shows that stylistic perturbations can compress or reweight margins, but they do not usually erase the fingerprint signal.

\paragraph{Contrastive step probe.} The baseline forced-completion probe establishes that models can harbor distinct procedural preferences, but leaves open the possibility that these preferences merely reflect superficial stylistic biases over actual computational grounding. Therefore, to test deeper method grounding, we introduce a contrastive step probe. Here, we score a correct step versus a plausible incorrect intermediate step that is matched in form for each heuristic. Table~\ref{tab:contrastive-method} illustrates. 

\begin{table}[htb]
    \centering
    \small
    \resizebox{\columnwidth}{!}{%
    \begin{tikzpicture}[
        font=\scriptsize,
        box/.style={
          rectangle,
          rounded corners=3pt,
          draw=black!55,
          fill=black!3,
          align=left,
          inner sep=5pt,
          text width=3.0cm
        },
        metric/.style={
          draw=blue!50!black,
          fill=blue!7,
          align=left,
          inner sep=5pt,
          text width=8.4cm
        },
        arrow/.style={-{Stealth[length=2mm]}, semithick, black!65},
        note/.style={font=\scriptsize\itshape, align=center}
      ]

      \node[box, minimum height=1.7cm] (item) {
        \textbf{1. Fix the same problem}\\
        One HDS item with the same operands $(a,b)$,
        shown either as text or as a rendered image.\\[2pt]
        Example: $47 \times 36$
      };

      \node[box, text width=4.9cm, minimum height=1.7cm, right=0.55cm of item] (pair) {
        \textbf{2. Build a matched step pair}\\
        Choose one heuristic (OT/DD/RC), then write two
        candidate next steps with identical wording.\\[2pt]
        \textcolor{green!40!black}{Correct:} $40 \times 36 + 7 \times 36 = 1440 + 252$\\
        \textcolor{red!55!black}{Incorrect:} $40 \times 36 + 7 \times 36 = 1440 + 262$
      };

      \node[box, minimum height=1.7cm, right=0.55cm of pair] (score) {
        \textbf{3. Score both continuations}\\
        Keep the prompt context fixed and compute
        forced-completion loss on the step tokens only:\\[2pt]
        $\ell_{\mathrm{correct}}$ and $\ell_{\mathrm{incorrect}}$
      };

      \node[metric, minimum width=8.4cm, below=0.8cm of pair] (agg) {
        \textbf{4. Aggregate the comparison}\\
        Preference event: $\ell_{\mathrm{correct}} < \ell_{\mathrm{incorrect}}$
        \hspace{0.7em}
        Loss gap: $\ell_{\mathrm{incorrect}} - \ell_{\mathrm{correct}}$\\
        Report both overall and separately on OT-, DD-, and RC-targeted items.
      };

      \draw[arrow] (item) -- (pair);
      \draw[arrow] (pair) -- (score);
      \draw[arrow] (pair.south) -- (agg.north);

      \node[note, above=0.15cm of pair] {
        Only arithmetic correctness changes; wording and local format are held fixed.
      };
    \end{tikzpicture}%
    }
    \caption{
    Contrastive step probe for text and image inputs, where we compare a correct and a plausible incorrect intermediate step with matched surface form; lower forced-completion loss indicates which step is more compatible with the model's continuation distribution.
    }
    \label{tab:contrastive-method}
  \end{table}

Most correct-step preferences are at ceiling (100\%) on target-aligned HDS items, so the probe primarily confirms basic grounding. The non-trivial signal comes from the loss gap between correct and incorrect. Although overall preference rates differ little by modality (text overall: \ContrastivePrefRateThirtyB{} / \ContrastivePrefRateTwoThirtyFiveB{}; image overall: \ContrastivePrefRateImageThirtyB{} / \ContrastivePrefRateImageTwoThirtyFiveB{}), they have mean loss gaps of \ContrastiveDeltaThirtyB{} / \ContrastiveDeltaTwoThirtyFiveB{} (text) and \ContrastiveDeltaImageThirtyB{} / \ContrastiveDeltaImageTwoThirtyFiveB{} (image) for correct and incorrect steps. Margins vary by heuristic, with loss-gap patterns varying also by model and modality (text OT: \ContrastiveOTDeltaThirtyB{} / \ContrastiveOTDeltaTwoThirtyFiveB{} vs DD: \ContrastiveDDDeltaThirtyB{} / \ContrastiveDDDeltaTwoThirtyFiveB{}; image OT: \ContrastiveOTDeltaImageThirtyB{} / \ContrastiveOTDeltaImageTwoThirtyFiveB{} vs DD: \ContrastiveDDDeltaImageThirtyB{} / \ContrastiveDDDeltaImageTwoThirtyFiveB{}). Table~\ref{tab:contrastive-results} summarizes.

These modality differences suggest that preferred arithmetic continuations are determined not solely by the mathematical structure of the problem but also at least to some extent by how the problem is presented. One possible explanation is that rendered equations change the salience of magnitude information \citep{dehaene2011number}, although our results cannot directly test this mechanism. See \S\ref{s:Traps} for an analysis of adversarial cases.
 
\subsection{LoRA Nudges Change Behavior, Do Not Improve Performance}

Given that the probes reveal stable heuristic preferences, we next ask whether we can \emph{actively} steer strategy selection with lightweight adaptation. Specifically, for each base model (Qwen3-VL-30B and Qwen3-VL-235B), we first trained three separate LoRA adapters on approximately 1,000 \emph{synthetic} heuristic-specific reasoning traces. These traces are generated programmatically with no overlap with the held-out HDS/Traps evaluation problems: an RC adapter on near-base rounding/compensation examples, a DD adapter on decomposition-friendly partial-product examples, and an OT adapter on carry-heavy column-style examples. These adapters were trained with early stopping on a validation split. 

We then evaluated the resulting adapters on the held-out HDS test split ($n=\NudgeTestCountThirtyB$ / $\NudgeTestCountTwoThirtyFiveB$) to ask whether forcing one trace family at inference time improves arithmetic behavior out of distribution. Applying the \NudgeLoRACountThirtyB{} / \NudgeLoRACountTwoThirtyFiveB{} trained heuristic LoRAs (RC, DD, OT) produced \NudgeTotalFlipsThirtyB{} / \NudgeTotalFlipsTwoThirtyFiveB{} behavioral (correctness) flips across \NudgeTotalComparisonsThirtyB{} / \NudgeTotalComparisonsTwoThirtyFiveB{} comparisons (\NudgeLoRACountThirtyB{} / \NudgeLoRACountTwoThirtyFiveB{} LoRAs $\times$ \NudgeTestCountThirtyB{} / \NudgeTestCountTwoThirtyFiveB{} problems per model)---\NudgeImprovedThirtyB{} / \NudgeImprovedTwoThirtyFiveB{} improved and \NudgeDegradedThirtyB{} / \NudgeDegradedTwoThirtyFiveB{} degraded---and \NudgeDetectionFlipsThirtyB{} / \NudgeDetectionFlipsTwoThirtyFiveB{} heuristic flips (changes in the lowest-loss template). 

Degradations tend to be more prevalent than improvements for both model sizes. A post-hoc answer analysis shows recurring partial-product omissions; magnitude slips and carry-drop errors were not observed (\NudgeDegradedPartialProductThirtyB{}, \NudgeDegradedMagnitudeSlipThirtyB{}, \NudgeDegradedCarryDropThirtyB{} for 30B; \NudgeDegradedPartialProductTwoThirtyFiveB{}, \NudgeDegradedMagnitudeSlipTwoThirtyFiveB{}, \NudgeDegradedCarryDropTwoThirtyFiveB{} for 235B). These results suggest that while LoRAs successfully learn heuristic-specific reasoning patterns (as evidenced by low validation loss), the model's internal heuristic router is better optimized for performance than any single heuristic; internal preferences appear to handle arithmetic reasoning better than any single enforced trace distribution.

To isolate whether observed degradations stem from heuristic specialization or from a more general disruption of the base model's native arithmetic routing by synthetic traces, we also evaluated a style-only control adapter trained solely to imitate reasoning-trace formatting without any heuristic-specific arithmetic content. Here, the STYLE control yields almost exclusively degraded cases.
In other words, even an adapter trained to imitate reasoning-trace format without imposing heuristic-specific arithmetic content still reduces correctness much more often than it helps. This suggests that the degradation seen under heuristic LoRAs is not just a consequence of specializing to OT, DD, or RC in particular; more broadly, pushing the model toward a synthetic trace distribution appears to disrupt the base model's native routing over arithmetic behaviors, at least in the training regime here.

\subsubsection{Heuristics in Update Space}

Finally, Table~\ref{tab:correlation} shows the cosine similarity between the effective LoRA update matrices for the three heuristic adapters; values are near zero. In sensitivity analysis, we retrain adapters for the same heuristic using different random seeds. Same-heuristic reruns show higher cosine similarity than cross-heuristic pairs (\EffectiveUpdateSameSeedAvgThirtyB{} vs.\ \EffectiveUpdatePrimaryCrossAvgThirtyB{} with gap \EffectiveUpdateSeedGapThirtyB{} for 30B; \EffectiveUpdateSameSeedAvgTwoThirtyFiveB{} vs.\ \EffectiveUpdatePrimaryCrossAvgTwoThirtyFiveB{} with gap \EffectiveUpdateSeedGapTwoThirtyFiveB{} for 235B), suggesting that the observed heuristic separation in parameter space is not primarily a high-dimensional artifact of LoRA training. Heuristic fine-tuning does not seem to trigger very similar updates in parameter space.

\section{Limitations}

Overall, our results show that exact multiplication performance in LLMs is governed by the computational burden induced by digit structure, secondarily modulated by input modality. Across text, image, and audio renderings, accuracy declines sharply with arithmetic load; forced-completion probes further suggest a preference for decomposition-style continuations whose margins shift under non-text modalities. Notwithstanding these results, our study has several limitations.

First, the task scope is narrow: we focus on multiplication, so patterns may differ for addition, division, symbolic algebra, or multi-step word problems. 

Second, model coverage remains limited. The multimodal accuracy analysis covers several model families (Table \ref{Tab:BigComparison}), but token-level fingerprinting and LoRA nudges are limited to Qwen3-VL-30B and Qwen3-VL-235B. 

Third, we rely on synthetic digit templates that may not match the distributions of real-world mathematical problems. The forced-preamble technique measures compatibility with heuristic-conditioned continuations, rather than a direct readout of the model's internal algorithm. 

Finally, our multimodal conditions use controlled renderings and do not encompass the full range of messy real-world inputs. This design is useful for isolating arithmetic structure and modality effects, but future work should test scanned documents, handwritten equations, spreadsheet screenshots, and tool-augmented agent settings. 

Regarding risks, while our benchmark and probing methodology are intended only for diagnostic characterization, the same forced-completion loss and LoRA-based techniques could in principle be repurposed to systematically map and exploit arithmetic failure modes in deployed multimodal systems, underscoring the importance of robust adversarial testing and verifiable computational safeguards. \hfill $\square$

\bibliography{custom}

\appendix

\section{Benchmark Construction Details and Evaluation Protocol}
\label{sec:protocol}

\paragraph{Decoding parameters.} All generations in the LoRA pipeline use temperature $= 0$ (deterministic). The maximum generation budget is 2048 tokens. LoRA learning rate and other hyper-parameters set using Tinker API defaults. We attempted to deploy an audio analysis on a Qwen3-omni variant but experienced technical difficulties. 

\paragraph{Uncertainty estimates.} Standard errors (±SE) are computed across problem instances. For accuracy and resolved-probe coverage, we use the binomial SE $\sqrt{p(1-p)/n}$ (reported in percentage points). For loss averages and resolved-probe target support, we use the sample standard deviation of per-problem values divided by $\sqrt{n}$, with target-support SE computed only across the resolved rows for that family. With deterministic decoding (temperature $=0$), these SEs reflect item-level variability.

\paragraph{HDS construction.} 
We construct the Heuristic-Disagreement Set (HDS) by seeding it with a hand-curated pool of problems that distinctively favor one of the three target heuristics. We then scale to a target count by sampling additional items into three roughly equal buckets based on design families: RC draws offsets within $\pm 10$ around bases $\{25, 50, 100, 200, 250, 500\}$; DD draws problems with trailing zeros, multiples of 25, or clean tens decompositions; and OT draws carry-heavy high-digit pairs or generic pairs explicitly away from near-base and zero cues. We also include small perturbation pairs that flip a single cue (e.g., near-base vs.\ not) to test the sensitivity of heuristic detection.

To formalize the target labels, each item is evaluated using a cost proxy based on primitive digit operations. OT uses the schoolbook cost of $n \cdot m$ digit multiplications plus a small carry/accumulation penalty; DD uses a one-sided expansion cost of $\min(m \cdot s, n \cdot t)$ plus additions for combining partial products, with explicit low-cost special cases for trailing-zero and quarter-hundred factors; and RC uses a shared-base correction cost built from the nearest round base and the two operand offsets. Items are retained in the final dataset only when the minimum-cost heuristic is uniquely separated from the runner-up by a fixed margin. Finally, we enforce problem uniqueness up to commutativity, generate adversarial traps disjoint from the HDS, and stratify train/val/test splits using \DatasetSplitRatios{} to preserve the intended target mix.

\paragraph{AI use statement.} AI was used for code correctness and grammar review.

\section{Notes on Operation Counts and Probes}
\label{sec:formal-notes}

\paragraph{Operation-count proxy.}
Let $a$ and $b$ be positive integers with digit lengths $n,m$ and non-zero digit counts $s,t$, respectively, so $1\le s\le n$ and $1\le t\le m$. A full digit expansion contains exactly $s t$ non-zero digit products. If these shifted product terms are summed sequentially, this requires at most $s t-1$ term additions, not counting the lower-level digit additions and carry propagation inside each addition. A one-sided digit expansion that expands the $s$ non-zero digits of $a$ against all $m$ digits of $b$ uses $m s$ digit multiplications, including multiplications by zero; exactly $s t$ of these products are non-zero. Similarly, expanding $b$ against $a$ uses $n t$ digit multiplications.

Now, the proxy
\[
    C=(n+m)(s+t)
\]
is monotone in total digit length and total non-zero digit count. It directly upper-bounds $m s$ and $n t$, since both appear as nonnegative terms in the expansion:
\[
    C = ns + nt + ms + mt.
\]
It also satisfies (by the arithmetic mean-geometric mean inequality):
\[
    st \le \frac{(s+t)^2}{4} \le \frac{(n+m)(s+t)}{4} = \frac{C}{4},
\]
where the second inequality uses $s+t\le n+m$. Thus, $C/4$ upper-bounds the number of non-zero pairwise digit products. $C$ is therefore only a proxy for intermediate computation number: it ignores carry propagation, aggregation order, and heuristic-specific shortcuts, and therefore does not, in general, preserve the exact ordering of operation counts across problems.

\paragraph{Likelihood-ratio probe.} Let $x_h$ be the heuristic template tokens and $\ell(h)=-\frac{1}{T_h}\sum_{t=1}^{T_h} \log p(x_{h,t}\mid c)$ the length-normalized cross-entropy under the same base model for context $c$ (the multiplication problem), computed on forced continuation tokens. If templates are token-length matched ($T_h=T$), then $-T \Delta \ell(h)$ equals the log-likelihood ratio $\log \frac{p(x_h\mid c)}{p(x_0\mid c)}$ between candidate templates under the same model. Selecting the minimum-loss heuristic is therefore the maximum-likelihood choice among candidates under equal priors; when token lengths differ, $\Delta \ell$ is a normalized naturalness score rather than an exact decision rule.

\section{Forced-Completion Probe Details and Results}
\subsection{Templates}
\label{sec:templates}

Our forced-completion loss probe uses heuristic-specific assistant continuations that are short and style-matched. The balanced template bank contains multiple paraphrases per heuristic; below, we show one representative continuation for each class:

\vspace{0.25cm}

\noindent \textbf{Columnar (OT)}: ``Column method: start with the ones digits [...]''

\vspace{0.25cm}

\noindent \textbf{Decomposition (DD)}: ``Decomposition: split one factor into place-value parts [...]''

\vspace{0.25cm}

\noindent \textbf{Rounding-Compensation (RC)}: ``Round and adjust: use a nearby round base, then compensate [...]''

\vspace{0.25cm}

\noindent \textbf{Neutral Baseline}: ``Let me solve this multiplication problem step by step [...]''

\vspace{0.25cm}

We compute cross-entropy loss over the template continuation tokens under forced completion and report the difference ($\Delta$loss) relative to the neutral baseline. The active bank is intentionally short and stylistically matched but not strictly token-length-identical, so we use a length-normalized loss to mitigate residual token-length differences.

\subsection{Forced-Completion Loss Results}
\label{sec:raw-loss}

For completeness, we report raw cross-entropy losses (with SE) under forced completion and the original modality comparison table. These absolute losses underlie the $\Delta$loss summaries in the main text.

\begin{table}[htb]
  \centering
  \begin{tabular}{lc}
    \hline
    Heuristic & 30B/235B Avg. Loss $\pm$ SE \\
    \hline
    DD (Decomposition) & \HDSTestPerpDDThirtyB{} $\pm$ \HDSTestPerpDDSEThirtyB{} / \HDSTestPerpDDTwoThirtyFiveB{} $\pm$ \HDSTestPerpDDSETwoThirtyFiveB{} \\
    RC (Rounding) & \HDSTestPerpRCThirtyB{} $\pm$ \HDSTestPerpRCSEThirtyB{} / \HDSTestPerpRCTwoThirtyFiveB{} $\pm$ \HDSTestPerpRCSETwoThirtyFiveB{} \\
    OT (Columnar) & \HDSTestPerpOTThirtyB{} $\pm$ \HDSTestPerpOTSEThirtyB{} / \HDSTestPerpOTTwoThirtyFiveB{} $\pm$ \HDSTestPerpOTSETwoThirtyFiveB{} \\
    \hline
  \end{tabular}
  \caption{Raw cross-entropy loss by heuristic template on the held-out probe split ($n=\HDSProbeCount$). Lower values indicate lower forced-completion loss (a deeper continuation preference).
  }
  \label{tab:loss-raw}
\end{table}

  \begin{table*}[htb]
    \centering
    \small
    \begin{tabular}{lcccc}
      \toprule
      Split & DD & RC & OT & Lowest-loss heuristic \\
      \midrule
      HDS (Test)
        & \HDSTestPerpDDThirtyB{} $\pm$ \HDSTestPerpDDSEThirtyB{}
        & \HDSTestPerpRCThirtyB{} $\pm$ \HDSTestPerpRCSEThirtyB{}
        & \HDSTestPerpOTThirtyB{} $\pm$ \HDSTestPerpOTSEThirtyB{}
        & DD \\
      Traps
        & \TrapsPerpDDThirtyB{} $\pm$ \TrapsPerpDDSEThirtyB{}
        & \TrapsPerpRCThirtyB{} $\pm$ \TrapsPerpRCSEThirtyB{}
        & \TrapsPerpOTThirtyB{} $\pm$ \TrapsPerpOTSEThirtyB{}
        & RC \\
      \bottomrule
    \end{tabular}
    \caption{Results for Qwen3-VL-30B-A3B. Entries are average forced-completion loss $\pm$ SE; lower values indicate deeper alignment.}
    \label{tab:loss-comparison-raw}
  \end{table*}

\section{Representative Analysis Examples}

\subsection{LoRA Training Examples}
\label{sec:training-examples}

We trained heuristic-specific LoRA adapters using synthetic reasoning traces. Below are representative examples showing the expected reasoning pattern for each heuristic. These traces were generated programmatically to ensure consistent structure while varying the specific realizations. 

\paragraph{Trace generation.}
Each heuristic uses a distinct algorithmic template:
\begin{itemize}[leftmargin=*]
    \item \textbf{RC}: For symmetric pairs ($a = \text{base} + k$, $b = \text{base} - k$), applies the difference of squares identity $(a)(b) = \text{base}^2 - k^2$. Non-symmetric pairs use a general rounding approach with four adjustment terms.
    \item \textbf{DD}: Decomposes the larger operand into tens and ones components, computes two partial products via the distributive property, and sums them.
    \item \textbf{OT}: Simulates columnar multiplication with explicit digit-by-digit products, carries, and partial product accumulation.
\end{itemize}

\paragraph{Problem selection.}
Problems were selected to match each heuristic's strengths, using programmatic generators used in the analysis. RC examples draw from bases \{25, 50, 75, 100, 125, 150, 175, 200, 250, 300, 400, 500\} with offsets 1--5, including both symmetric and mixed near-base pairs. DD examples mix trailing-zero factors, easy tens-plus-ones decompositions, and generic two-digit pairs. OT examples mix carry-heavy 2--3 digit pairs with generic multi-digit numbers. The generator also supports a STYLE control dataset with generic trace formatting but no heuristic-specific intermediate computation. To ensure unbiased evaluation, we exclude all HDS validation/test problems, Traps problems, and held-out multimodal problems from training data.

% Auto-generated LoRA training examples
% Generated by Scripts/analysis/GenerateResultsFigures.py

% Requires: \usepackage{listings} in preamble

\subsection*{Rounding-Compensation (RC) Example}
\textsc{Problem:} $399 \times 399 = 159201$

\begin{lstlisting}[basicstyle=\small\ttfamily,breaklines=true,literate={×}{{$\times$}}1 {²}{{$^2$}}1]
What is 399 × 399?
Let me round to convenient bases and adjust.
399 is close to 400 (difference: -1).
399 is close to 400 (difference: -1).
Start with 400 × 400 = 160000.
Adjustment for 399: 400 × -1 = -400.
Adjustment for 399: -1 × 400 = -400.
Cross term: -1 × -1 = +1.
Total: 160000 + (-400) + (-400) + (+1) = 159201.
Answer: 159201
\end{lstlisting}

\subsection*{Decomposition (DD) Example}
\textsc{Problem:} $99 \times 40 = 3960$

\begin{lstlisting}[basicstyle=\small\ttfamily,breaklines=true,literate={×}{{$\times$}}1 {²}{{$^2$}}1]
What is 99 × 40?
Let me decompose 99 into 90 + 9.
First compute 90 × 40:
90 × 40 = 3600.
Then compute 9 × 40:
9 × 40 = 360.
Now sum the partial products:
3600 + 360 = 3960.
Answer: 3960
\end{lstlisting}

\subsection*{Ones-Then-Tens (OT) Example}
\textsc{Problem:} $79 \times 78 = 6162$

\begin{lstlisting}[basicstyle=\small\ttfamily,breaklines=true,literate={×}{{$\times$}}1 {²}{{$^2$}}1]
What is 79 × 78?
Let me use column multiplication step by step.
Step 1: Multiply 79 by ones digit 8:
  9 × 8 = 72, write 2, carry 7.
  7 × 8 = 56, plus carry = 63.
  First partial product: 632.
Step 2: Multiply 79 by tens digit 7:
  9 × 7 = 63, write 3, carry 6.
  7 × 7 = 49, plus carry = 55.
  Second partial product: 553 (shifted by 10 = 5530).
Step 3: Add partial products:
  632 + 5530 = 6162.
Answer: 6162
\end{lstlisting}

\section{Adapter Weight Similarity}
\label{sec:similarity}

Table \ref{tab:correlation} shows the cosine similarity of the effective LoRA updates ($\Delta W = BA$) between the three heuristic adapters (OT, DD, and RC) for both the 30B and 235B models.

\begin{table}[htb]
  \centering
  \small
  \begin{tabular}{lccc}
    \hline
    \multicolumn{4}{c}{\it Qwen3-VL-30B} \\
    \hline
    & \textbf{OT} & \textbf{DD} & \textbf{RC} \\
    \hline
    OT & 1.00 & $\CorrOTDDThirtyB{}$ & \CorrOTRCThirtyB{} \\
    DD & $\CorrOTDDThirtyB{}$ & 1.00 & $\CorrDDRCThirtyB{}$ \\
    RC & \CorrOTRCThirtyB{} & $\CorrDDRCThirtyB{}$ & 1.00 \\
    \hline
    \multicolumn{4}{c}{\it Qwen3-VL-235B} \\
    \hline
    & \textbf{OT} & \textbf{DD} & \textbf{RC} \\
    \hline
    OT & 1.00 & $\CorrOTDDTwoThirtyFiveB{}$ & \CorrOTRCTwoThirtyFiveB{} \\
    DD & $\CorrOTDDTwoThirtyFiveB{}$ & 1.00 & $\CorrDDRCTwoThirtyFiveB{}$ \\
    RC & \CorrOTRCTwoThirtyFiveB{} & $\CorrDDRCTwoThirtyFiveB{}$ & 1.00 \\
    \hline
  \end{tabular}
  \caption{
  Cosine similarity of effective LoRA updates ($\Delta W = BA$) between the three heuristic adapters (OT/DD/RC) for 30B and 235B ($\rho$ denotes cosine similarity). OT and DD updates are nearly orthogonal ($\rho = \CorrOTDDThirtyB{}$ / \CorrOTDDTwoThirtyFiveB{}), suggesting the model recruits distinct parameter subspaces to implement these strategies.
  }
  \label{tab:correlation}
\end{table}

% Auto-generated merged template-variability appendix
% Generated by Scripts/analysis/GenerateResultsFigures.py

\begin{table}[htbp]
\centering \small 
\caption{Template-variability robustness summary by model (mean within-problem standard deviation across paraphrase losses)}
\label{tab:template-variability}
\begin{tabular}{@{}lcccc@{}}
\toprule
Profile & DD std & OT std & RC std & Mean std \\
\midrule
\multicolumn{5}{@{}l@{}}{\textbf{30B}} \\
Balanced Text & 0.4402 & 0.4631 & 0.2750 & 0.3928 \\
Style-Mismatch Text & 1.0000 & 0.7365 & 0.4042 & 0.7136 \\
Balanced Image & 0.3767 & 0.6300 & 0.1397 & 0.3821 \\
Style-Mismatch Image & 0.7704 & 0.6025 & 0.6218 & 0.6649 \\
\midrule
\multicolumn{5}{@{}l@{}}{\textbf{235B}} \\
Balanced Text & 0.5365 & 0.4384 & 0.2213 & 0.3987 \\
Style-Mismatch Text & 0.8869 & 1.0133 & 0.2363 & 0.7122 \\
Balanced Image & 0.5461 & 0.4302 & 0.3785 & 0.4516 \\
Style-Mismatch Image & 0.6440 & 0.9168 & 0.4381 & 0.6663 \\
\bottomrule
\end{tabular}
\end{table}

\subsection{Adversarial Traps}\label{s:Traps}

To examine robustness in adversarial cases, we also study \emph{adversarial traps}: test-only problems crafted to target known weaknesses of specific heuristics, serving as a focused probe of strategy robustness beyond the HDS setting. In our trap set ($n=\TrapsCount{}$), the analysis contributes a direct comparison of heuristic preference signals under adversarial pressure, clarifying which strategies degrade and which remain stable relative to the held-out probe split.

When presented with anti-round traps, the smaller 30B model shifts toward RC: its RC target-support share rises from $26.5\% \pm 0.5\%$ on the standard held-out split to $34.5\% \pm 1.4\%$ on the traps. The 235B model also increases on RC-targeted items, but much more modestly, moving from $12.4\% \pm 0.3\%$ to $15.6\% \pm 0.3\%$. We observe different dynamics with other strategies: DD target support moves from $27.9\% \pm 0.3\%$ to $24.1\% \pm 0.4\%$ for 30B and from $15.2\% \pm 0.2\%$ to $19.3\% \pm 0.2\%$ for 235B on missing-term traps, while OT target support moves from $25.8\% \pm 0.4\%$ to $31.7\% \pm 1.4\%$ for 30B and from $5.9\% \pm 0.3\%$ to $12.1\% \pm 0.3\%$ for 235B on OT-targeted traps. These results indicate that adversarial operand cues redistribute heuristic preference mass, with the strongest RC-specific shift appearing in the smaller model.

\begin{table*}[htb]
  \centering\small
  \begin{tabular}{llcc}
    \hline
    \textbf{Metric} & \textbf{Model} & \textbf{Text} & \textbf{Image} \\
    \hline
    Overall preference for correct step & 30B & \ContrastivePrefRateThirtyB{} $\pm$ \ContrastivePrefRateSEThirtyB{} & \ContrastivePrefRateImageThirtyB{} $\pm$ \ContrastivePrefRateSEImageThirtyB{} \\
    & 235B & \ContrastivePrefRateTwoThirtyFiveB{} $\pm$ \ContrastivePrefRateSETwoThirtyFiveB{} & \ContrastivePrefRateImageTwoThirtyFiveB{} $\pm$ \ContrastivePrefRateSEImageTwoThirtyFiveB{} \\
    Overall loss gap (incorrect - correct) & 30B & \ContrastiveDeltaThirtyB{} $\pm$ \ContrastiveDeltaSEThirtyB{} & \ContrastiveDeltaImageThirtyB{} $\pm$ \ContrastiveDeltaSEImageThirtyB{} \\
    & 235B & \ContrastiveDeltaTwoThirtyFiveB{} $\pm$ \ContrastiveDeltaSETwoThirtyFiveB{} & \ContrastiveDeltaImageTwoThirtyFiveB{} $\pm$ \ContrastiveDeltaSEImageTwoThirtyFiveB{} \\
    \hline
    \multicolumn{4}{l}{\textit{By target heuristic (preference rate):}} \\
    \quad DD preference for correct step (DD-targeted items) & 30B & \ContrastiveDDPrefRateThirtyB{} $\pm$ \ContrastiveDDPrefRateSEThirtyB{} & \ContrastiveDDPrefRateImageThirtyB{} $\pm$ \ContrastiveDDPrefRateSEImageThirtyB{} \\
    & 235B & \ContrastiveDDPrefRateTwoThirtyFiveB{} $\pm$ \ContrastiveDDPrefRateSETwoThirtyFiveB{} & \ContrastiveDDPrefRateImageTwoThirtyFiveB{} $\pm$ \ContrastiveDDPrefRateSEImageTwoThirtyFiveB{} \\
    \quad RC preference for correct step (RC-targeted items) & 30B & \ContrastiveRCPrefRateThirtyB{} $\pm$ \ContrastiveRCPrefRateSEThirtyB{} & \ContrastiveRCPrefRateImageThirtyB{} $\pm$ \ContrastiveRCPrefRateSEImageThirtyB{} \\
    & 235B & \ContrastiveRCPrefRateTwoThirtyFiveB{} $\pm$ \ContrastiveRCPrefRateSETwoThirtyFiveB{} & \ContrastiveRCPrefRateImageTwoThirtyFiveB{} $\pm$ \ContrastiveRCPrefRateSEImageTwoThirtyFiveB{} \\
    \quad OT preference for correct step (OT-targeted items) & 30B & \ContrastiveOTPrefRateThirtyB{} $\pm$ \ContrastiveOTPrefRateSEThirtyB{} & \ContrastiveOTPrefRateImageThirtyB{} $\pm$ \ContrastiveOTPrefRateSEImageThirtyB{} \\
    & 235B & \ContrastiveOTPrefRateTwoThirtyFiveB{} $\pm$ \ContrastiveOTPrefRateSETwoThirtyFiveB{} & \ContrastiveOTPrefRateImageTwoThirtyFiveB{} $\pm$ \ContrastiveOTPrefRateSEImageTwoThirtyFiveB{} \\
    \hline
    \multicolumn{4}{l}{\textit{By target heuristic (loss gap):}} \\
    \quad DD loss gap (DD-targeted items) & 30B & \ContrastiveDDDeltaThirtyB{} $\pm$ \ContrastiveDDDeltaSEThirtyB{} & \ContrastiveDDDeltaImageThirtyB{} $\pm$ \ContrastiveDDDeltaSEImageThirtyB{} \\
    & 235B & \ContrastiveDDDeltaTwoThirtyFiveB{} $\pm$ \ContrastiveDDDeltaSETwoThirtyFiveB{} & \ContrastiveDDDeltaImageTwoThirtyFiveB{} $\pm$ \ContrastiveDDDeltaSEImageTwoThirtyFiveB{} \\
    \quad RC loss gap (RC-targeted items) & 30B & \ContrastiveRCDeltaThirtyB{} $\pm$ \ContrastiveRCDeltaSEThirtyB{} & \ContrastiveRCDeltaImageThirtyB{} $\pm$ \ContrastiveRCDeltaSEImageThirtyB{} \\
    & 235B & \ContrastiveRCDeltaTwoThirtyFiveB{} $\pm$ \ContrastiveRCDeltaSETwoThirtyFiveB{} & \ContrastiveRCDeltaImageTwoThirtyFiveB{} $\pm$ \ContrastiveRCDeltaSEImageTwoThirtyFiveB{} \\
    \quad OT loss gap (OT-targeted items) & 30B & \ContrastiveOTDeltaThirtyB{} $\pm$ \ContrastiveOTDeltaSEThirtyB{} & \ContrastiveOTDeltaImageThirtyB{} $\pm$ \ContrastiveOTDeltaSEImageThirtyB{} \\
    & 235B & \ContrastiveOTDeltaTwoThirtyFiveB{} $\pm$ \ContrastiveOTDeltaSETwoThirtyFiveB{} & \ContrastiveOTDeltaImageTwoThirtyFiveB{} $\pm$ \ContrastiveOTDeltaSEImageTwoThirtyFiveB{} \\
    \hline
  \end{tabular}
  \caption{Contrastive step probe results on HDS test problems ($n=\ContrastiveCountThirtyB{}$). Preference rates report the share of items where the correct step has lower loss; loss gaps are mean incorrect minus correct loss.}
  \label{tab:contrastive-results}
\end{table*}

\section*{Data Availability}

We release the multimodal benchmarks across text, image, and audio renderings of the same multiplication items at:
\begin{itemize}
\item[] \url{https://huggingface.co/datasets/cjerzak/MultimodalMathBenchmarks}
\end{itemize} 

\end{document}